\documentclass[compsoc,conference,a4paper,10pt,times]{IEEEtran}

\IEEEoverridecommandlockouts

\usepackage{tikz}
\usepackage{cite}
\usepackage[numbers]{natbib}
\usepackage{amsmath,amssymb,amsfonts}
\usepackage{algorithmic}
\usepackage{graphicx}
\usepackage{textcomp}
\usepackage{bmpsize}
\usepackage{xcolor}
\usepackage{lipsum}
\usepackage[colorlinks=true,urlcolor=black]{hyperref}
\usepackage{tabularx}
\usepackage{multirow,multicol,enumitem,epsfig,url,makecell}
\usepackage{mathtools}
\usepackage{subcaption}
\usepackage{pgfplots}
\DeclareUnicodeCharacter{2212}{−}
\usepgfplotslibrary{groupplots,dateplot}
\usetikzlibrary{patterns,shapes.arrows}
\pgfplotsset{compat=newest}

\usepackage{booktabs}       
\usepgfplotslibrary{statistics} 
\tikzstyle{line}=[draw]

\newtheorem{definition}{Definition}
\newtheorem{adversary}{Attack}

\newtheorem{game}{Attack Game}

\newcommand{\paragraphbe}[1]{\vspace{0.08in} \noindent{\bf \em #1}. }

\newcommand{\D}{\mathcal{D}} 
\newcommand{\M}{M} 
\newcommand{\Ms}{M_{S}} 
\newcommand{\X}{\mathcal{X}} 
\newcommand{\Y}{\mathcal{Y}} 
\newcommand{\G}{\mathcal{G}} 
\newcommand{\Adv}{\mathcal{A}} 
\newcommand{\N}{\mathcal{N}} 
\newcommand{\FN}{\mbox{FN}} 
\newcommand{\train}{S} 
\newcommand{\A}{A} 

\def\compileFigures{0}
\newcommand{\filename}{main}
\newcounter{figureNumber}

\if\compileFigures1
	\usetikzlibrary{external}
\fi

\author{
	\IEEEauthorblockN{Hongyan Chang and Reza Shokri}
	\IEEEauthorblockA{\textit{Department of Computer Science, National University of Singapore (NUS)}\\firstname@comp.nus.edu.sg}
}

\title{On the Privacy Risks of Algorithmic Fairness}

\begin{document}

\maketitle

\begin{abstract}
	Algorithmic fairness and privacy are essential pillars of trustworthy machine learning.  Fair machine learning aims at minimizing discrimination against protected groups by, for example, imposing a constraint on models to equalize their behavior across different groups. This can subsequently change the {\em influence} of training data points on the fair model, in a disproportionate way.  We study how this can change the information leakage of the model about its training data.  We analyze the privacy risks of group fairness (e.g., equalized odds) through the lens of \emph{membership inference attacks}: inferring whether a data point is used for training a model.  We show that fairness comes at the cost of privacy, and this cost is not distributed equally: the information leakage of fair models increases significantly on the unprivileged subgroups, which are the ones for whom we need fair learning. We show that the more biased the training data is, the higher the privacy cost of achieving fairness for the unprivileged subgroups will be.  We provide comprehensive empirical analysis for general machine learning algorithms.
\end{abstract}



\section{Introduction}\label{sec:introduction}

Machine learning algorithms can be discriminatory against different groups, due to their training algorithm or the bias in their training data.  This is shown in various applications from computer vision~\cite{buolamwini2018gender} to word embedding~\cite{bolukbasi2016man}.  Fair machine learning aims at addressing this issue~\cite{agarwal2018reductions, calders2009building, dwork2012fairness, hardt2016equality, kamishima2011fairness, kusner2017counterfactual, madras2018learning, zafar2015fairness, zafar2017fairness, zemel2013learning}.  At the same time, machine learning models are shown to leak significant amount of information about their training data, which can be exploited by inference attacks~\cite{song2020information, shokri2019privacy, yeom2018privacy, song2019privacy, shokri2017membership, sablayrolles2019white, nasr2019comprehensive}.  Privacy-preserving machine learning aims at alleviating this problem~\cite{shokri2019privacy, abadi2016deep, cummings2019compatibility, chaudhuri2011differentially, papernot2018scalable}.

Privacy and fairness, as two societal concerns about machine learning, do not exist in isolation.  For instance, in the recidivism prediction application, demographic groups (e.g., black defendants and white defendants) should experience similar treatments, namely similar prediction accuracy. Simultaneously, participation in the training data means that the individual had once committed a crime, which is very sensitive and needs to be kept private.  Thus, fairness and privacy are both needed for an ethical use of machine learning.  It is, therefore, imperative to understand the interactions between them.

Training a model with privacy guarantees can lead to disparate accuracy across different groups in the population.  Notably, differentially private models (e.g., DP-SGD~\cite{abadi2016deep}) impose a larger accuracy reduction on ``under-represented'' subgroups~\cite{bagdasaryan2019differential}. In other words, privacy can come at the cost of fairness.  In this paper, we ask the related yet complementary question: \emph{Is there a privacy cost for achieving group fairness?}  We study if enforcing fairness constraints on the learning algorithm can impact its privacy risk with respect to the training data.

One way to address this question is through analyzing models which are trained with differential privacy and fairness constraints~\cite{ding2020differentially, xu2019achieving, cummings2019compatibility}, and evaluating the compatibility of the two measures.  In this paper, we choose a complementary adversarial approach.  We formalize privacy risk as the success of \emph{membership inference attacks} against machine learning models.  This reflects the information leakage of a model about the \emph{individual} data points in its training set.  We use this to quantify how fair and unconstrained models differ in their information leakage about different groups in their training data.

We assume the adversary observes the model's predictions and aims to find a ``distinguisher'' that identifies members of the training set from non-members.  Finding a single global  ``distinguisher'' for all data points (e.g., a threshold on the loss of the model on its inputs) is the prevalent method in the existing membership inference attacks~\cite{shokri2017membership, yeom2018privacy, sablayrolles2019white}.  However, this approach is evidently sub-optimal. In practice, data samples from different groups can have different underlying distributions.  Thus, the machine learning models might learn different patterns on each group, for the same task.  This is what leads to performance disparity of the model, which motivates using fair algorithms.  In other words, the way that a model's predictions (loss) on training versus test data differs, can be distinct from one group to another.  This can be exploited by the membership inference attacks.  We, thus, propose an effective attack strategy where the adversary finds a ``distinguisher'' per sub-group. We empirically show that this simple modification of existing membership inference attacks results in a higher attack accuracy, thus leads to a more accurate estimation of the privacy risk. We focus on the information leakage of models through their predictions: \emph{black box} setting, in which the adversary cannot observe the internal state of the model (e.g., its parameters and gradients).

We show that fairness comes at the cost of privacy.  Based on our attack strategy, we empirically show that the fairness-aware learning has a disparate impact on the privacy risk of subgroups, and in particular, it increases the privacy risk of the unprivileged subgroup.  Furthermore, there is a trade-off between fairness and privacy. When the underlying data and the corresponding unconstrained model are more ``unfair'', the trained fair models leak more information about the unprivileged subgroups.  Additionally, the more fair a model is, the higher the privacy risk of the model on the unprivileged subgroups will be.

Fairness constraints force models to perform equally on all the subgroups.  Yet, when the size of an unprivileged subgroup is small, or due to its complexities and variance, it is hard to fit a model on it, fair models memorize the training data from the unprivileged subgroups (instead of learning a general pattern on them).  This memorization gives rise to a high privacy risk, and it becomes easier for the adversary to infer membership of the training data. Hence, the unprivileged subgroups have a higher privacy risk on fair models.

We perform extensive empirical analysis on machine learning models.  In the evaluation, we use synthetic data to analyze how, when, and why fair models leak more information about their training data. We also conduct experiments on multiple real-world datasets, including the Law School dataset~\cite{lawdataset}, Bank Marketing dataset~\cite{ucidataset}, and COMPAS datasets~\cite{compasdataset}.  

\section{Background}\label{sec:background}

In this section, we present the definitions for group fairness and data privacy.  

\subsection{Machine Learning}\label{subsec:machine_learning}

We consider supervised machine learning with the focus on classification tasks.  Let ${\M: \X \to \Y}$ be a machine learning model that maps the input (feature) space ${\X}$ to the output (label) space ${\Y}$.  Let $\ell$ be the loss function, and $\train$ be the training set of size $n$.  We use $\Ms$ to denote the model trained on $\train$.  We assume each data point in $\train$ is sampled i.i.d. from a data distribution $\D$.  We refer to the models obtained by the standard learning algorithm (without fairness requirements) as \emph{unconstrained models}.

\subsection{Fairness}\label{subsec:fairness}

The central problem in fair machine learning is to ensure that the machine learning model does not discriminate against individuals with specific values in their protected attributes (e.g., race, gender).  We represent each data point as $z = (x,g,y) \in \mathcal{X}\times \mathcal{G}\times \mathcal{Y}$, where $x \in \X$ is the set of model's input features, $g \in \G$ is the \emph{protected attribute}, and $y \in \Y$ is the label.  The protected attributes partitions the population.  Let $\D_{g}^y$ denote the distribution of data with protected attribute $G = g$ and label $Y = y$.  We use $G_{g}$ to represent the group $\{(X,G,Y)|G=g\}$. We use $G_{g}^y$ to represent the subgroup with $\{(X,G,Y)|G=g, Y=y\}$.

Without loss of generality, we consider the binary classification setting, where $y \in \Y = \{-,+\}$.  We use $X$, $Y$, and $G$ to denote the random variables associated with the feature vector, the label, and the protected attribute, respectively. Input features $X$ might include $G$.  For instance, each data point can correspond to an applicant in a loan approval system, and $X$ could be the demographics, income level, and loan amount, and $G$ could be the applicant's race.  Input features $X$ might include $G$, or otherwise include other features such as zip code, which is often correlated with race.

\emph{Group fairness} measures require that different protected groups, on average, are treated similarly by the model.  In this paper, we mainly focus on \emph{equalized odds}, which is a widely-used definition for group fairness~\cite{hardt2016equality}. We also evaluate other group fairness notions, notably equal opportunity~\cite{hardt2016equality} and false-positive parity~\cite{hardt2016equality}, which are explained in Section~\ref{sec:evaluation}.

A model is fair with respect to equalized odds if the model error on different groups is the same.  In other words, given the true label for a data point, a fair model's prediction on a data point and its protected attribute should be conditionally independent.  Following prior work~\cite{agarwal2018reductions, donini2018empirical}, we use a relaxed notion of equalized odds:
\begin{definition}[$\delta$ - Equalized Odds Fairness] \label{def:eo_fairness}
	A classifier $\M$ satisfies $\delta$-Equalized Odds with respective to the protected attribute $\G$, if for all $g, g' \in \G$, the false positive rate and false negative rate of the classifier for group $\{G = g\}$ and $\{G = g' \}$ are within $\delta$ range of one another.
	\begin{align} \label{eq:eo-fairness}
		\Delta(\M, \D) \triangleq  \nonumber \\
		\max_{\substack{y\in\{-,+\} \\ g,g' \in \G}} \bigg | &\Pr_\D[\M(X)\neq y|S=g, Y=y] \nonumber \\
		 - &\Pr_\D[\M(X)\neq y|S=g', Y=y] \bigg | \le \delta ,
	\end{align}
	where the probabilities are computed over the data distribution $\D$.  We refer to $\Delta$ as the model's {\bf fairness gap} under equalized odds.  A model satisfies exact fairness under equalized odds when $\delta=0$.
\end{definition}

In practice, the data distribution $\D$ is unknown. A fair model is thus obtained by ensuring $\delta$-fairness empirically on the training set $\train$.  We can minimize the model's empirical loss under $\delta$-fairness as a constraint, or through post-processing~\cite{hardt2016equality}.  We refer to $\delta$ as the \emph{enforced fairness level}.

\subsection{Membership Privacy}

	We follow the privacy notion underlying \emph{differential privacy}~\cite{dwork2006calibrating}: privacy is preserved if the output distributions of an algorithm on two neighboring datasets are almost indistinguishable.  In other words, given an observation from a privacy-preserving algorithm, an adversary is unable to tell whether the record of a participant is in the input dataset or not.

	Thus, we measure the privacy risk of an individual as the success of an adversary whose goal is to infer whether the individual's record is part of the input dataset.  Such attacks are called \emph{membership inference attacks} which are used as a \emph{tool} to measure information leakage of different machine learning algorithms, including deep learning algorithms~\cite{shokri2017membership}, adversarially robust learning algorithms \cite{song2019privacy}, learning algorithms for explanations models~\cite{shokri2019privacy}, learning algorithms for embedding models~\cite{song2020information}, and reinforcement learning algorithms~\cite{pan2019you}.


\section{Privacy Analysis}\label{sec:problem_statement}

In this section, we present our generic approach for the analysis of privacy risks. In Section~\ref{sec:evaluation}, we use this framework to perform our empirical analysis.


\subsection{Definition of Privacy Risk}\label{subsec:privacy_risk}

We compute privacy risk as the accuracy of membership inference attacks (i.e., the probability that an adversary can correctly infer if a data point is part of the input dataset).  We use the following game between an adversary and a challenger, to formalize membership inference attacks. 

\begin{game}[Membership Inference]\label{exp:mem_attack_point}
	\begin{enumerate}
		\item Adversary chooses a data point $z$, and sends it to the challenger.  
		\item Challenger chooses a secret bit ${b \leftarrow \{0,1\}}$ uniformly at random, and samples dataset $S \sim \D^{n}$.  If $b=1$, the challenger overwrites a random element in $S$ with $z$. 
		\item Challenger runs algorithm $\A$ on $\train$ and sends its outputs~$\A_\train$ to the adversary.
		\item Adversary runs an inference attack $\Adv$, and tries to infer the secret bit as $\hat{b} \in \{0,1\}$.
		\item The game outputs $1$ (indicating that adversary wins) if $\hat{b} = b$, and $0$ otherwise.
	\end{enumerate}
\end{game}

We define privacy risk of algorithm $\A$ with respect to an individual data point $z$ as the probability that the most powerful adversary wins the attack game.  

\begin{definition}[Individual Privacy Risk]\label{def:membership_advantage}
	Given an algorithm $\A$ and data distribution $\D$, the privacy risk of $\A$ with respect to data point $z$ is
	\begin{equation}
		\mbox{PR}(z, \A, \D)  \triangleq \max_{\Adv} \Pr[ \mbox{Attack Game outputs } 1], \nonumber
	\end{equation}
	where the probability is taken over all the randomness in Attack Game~\ref{exp:mem_attack_point}. 
\end{definition}

The individual privacy risk is equivalent to the average true positive and true negative rates ${\frac{1}{2}(\Pr[ \hat{b}=1 | b=1] + \Pr[ \hat{b}=0 | b=0])}$ of the adversary.

\begin{definition}[Subgroup Privacy Risk]\label{def:subgroup_privacy}
	We define the privacy risk of algorithm $\A$ with respect to subgroup~$G_{g}^y$ (i.e., data points with label $y$ and protected attribute~$g$) as 
	\begin{equation}
		\mbox{PR}(G_{g}^y, \A, \D) \triangleq \mathbb{E}_{z \sim \D_{g}^y}[\mbox{PR}(z, \A, \D)], 
	\end{equation}
	which is the expectation of the privacy risk of individual data points in $G_{g}^y$.
\end{definition}


\subsection{Quantifying Privacy Risk}\label{subsec:method_for_measuring_privacy}

We assume the adversary has black-box access to the model, and can compute the loss of the model on any input data.  To run a membership inference attack in this setting, the adversary can use a single attack model on all input data~\cite{shokri2017membership}.  A simple attack model is to compare the model's loss on an input with a threshold.  The attack outputs ``member'' if the loss is below the threshold, and ``non-member'' otherwise~\cite{yeom2018privacy, sablayrolles2019white}.  This attack is based on a single distinguisher between members and non-members, which makes the attack sub-optimal.  In practice, the training data might be composed of samples from (slightly) different distributions.  This means the model might learn (and memorize) distinct patterns from different parts of the training set.  Thus, the membership inference attack that is adapted to each sub-population should potentially perform better than a fixed attack model (based on a single loss threshold).  Specifically, in the case of fair models, where the algorithm explicitly treats different subgroups differently in order to equalize its error across them, the adversary can design a separate membership inference attack for each subgroup. 

Based on this, we propose to use different loss thresholds for different subgroups.

\begin{adversary}\label{adv:multiple_loss}
	Let $\tau^{(g,y)}$ be the loss threshold for distinguishing training data members from non-members in subgroup~$G_{g}^y$. On an input $z = (x,g,y)$, and machine learning model $\A_S$, the adversary proceeds as follows:
	\begin{enumerate}
		\item Query the model to obtain $\ell(\A_S,z)$.
		\item Output $1$ (``member'') if $\ell(\A_S,z) < \tau^{(g,y)}$ and $0$ (``non-member'') otherwise.
	\end{enumerate}
\end{adversary}

In practice, the adversary can compute the loss threshold based on the knowledge about the population~\cite{sablayrolles2019white, yeom2018privacy} or through using shadow models~\cite{shokri2017membership}. In the evaluation, we find a loss threshold that best separates the members and non-members of each subgroup.  We measure the individual privacy risk and subgroup privacy risk based on Adversary~\ref{adv:multiple_loss}.  Note that the attacker can only obtain the best threshold when he knows the training set. Thus, measuring the privacy risks using the best loss threshold gives us a closer estimation of privacy risk under the strongest attack.


\section{Empirical Analysis}\label{sec:evaluation}

We use the reductions approach for training fair machine learning models~\cite{agarwal2018reductions}\footnote{\url{https://github.com/fairlearn/fairlearn}}. The algorithm produces a randomized classifier, and we compute its expected accuracy in our analysis. 

We analyze the success of adversary in Attack Game~\ref{exp:mem_attack_point}
to quantify privacy risk for individual data points and subgroups.  We assume Adversary~\ref{adv:multiple_loss} has black-box access to models, and can observe the model's loss on each query.  We measure the \textbf{privacy cost} of fair algorithms as the difference between the privacy risk of fair and unconstrained models.  



\begin{figure*}[t!]
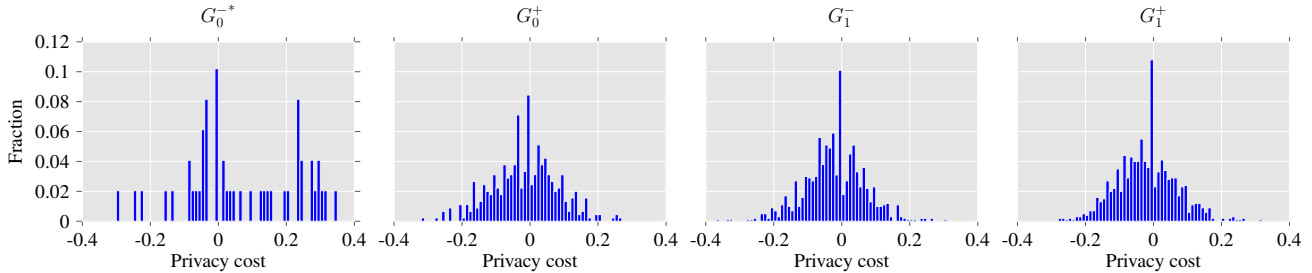

	\centering
	\if\compileFigures1
  \resizebox{\textwidth}{!}{\input{figure_scripts/fig_histogram_attack_accuracy.tex}}
	\else
	\resizebox{\textwidth}{!}{\includegraphics[]{fig/\filename-figure\thefigureNumber.pdf}}
	\stepcounter{figureNumber}
	\fi
  \caption{\small Histogram for individual privacy cost, across different subgroups, on models trained on synthetic data. The x-axis is the privacy cost, which is the difference between individual privacy risk on fair models and unconstrained models.  The average value of the individual privacy cost is $0.069$, $-0.015$, $-0.02$, $-0.02$ for subgroups $G_0^{-}$, $G_0^{+}$, $G_1^{-}$, and $G_1^{+}$ respectively.}
  \label{fig:privacy_gap_four_subgroups}
\end{figure*}


\begin{figure}[t!]
	\centering
	\if\compileFigures1
  \resizebox{0.8\columnwidth}{!}{
\begin{tikzpicture}
\large
\begin{axis}[
axis background/.style={fill=white!89.8039215686275!black},
axis line style={white},
legend cell align={left},
legend style={fill opacity=0.8, draw opacity=1, text opacity=1, at={(0.03,0.97)}, anchor=north west, fill=white!89.8039215686275!black},
tick align=outside,
tick pos=left,
legend pos = north west,
x grid style={white},
xlabel={Privacy risk},
xmajorgrids,
xmin=0.4, xmax=1,
xtick style={color=white!33.3333333333333!black},
y grid style={white},
ylabel={Memorization},
ymajorgrids,
ymin=-0.1, ymax=1.6,
ytick style={color=white!33.3333333333333!black}
]

\addlegendimage{blue, mark=o, mark size=2, only marks}
\addlegendentry{\small Unconstrained};
\addlegendimage{red, mark=*, mark size=2, only marks}
\addlegendentry{\small Fair ($\delta=0.001$)};

\addplot [semithick, blue, mark=triangle, mark size=3, mark options={solid}, only marks]
table {%
0.633333333333333 0.0656249046325683
0.625 0.105878844857216
0.526785714285714 0.276171142501491
0.583333333333333 0.247487164205975
0.722222222222222 0.124479942851596
0.53125 0.322569454354899
0.553571428571429 0.22321002345
0.6 0.0882550795873007
0.553571428571429 0.197146262973547
0.513392857142857 0.0285104098064559
0.566666666666667 0.747531032562256
0.589285714285714 0.487171614808696
0.595022624434389 0.201055306115301
0.479638009049774 0.130104608125816
0.62200956937799 0.124150093377492
0.625 0.667891972594791
0.5 0.538655108875699
0.5 0.553216019315583
};
\addplot [semithick, blue, mark=triangle, mark options={rotate=180}, mark size=3, only marks]
table {%
0.522321428571429 -0.0142464842647314
};

\addplot [semithick, blue, mark=pentagon, mark size=3, mark options={solid}, only marks]
table {%
0.678571428571429 0.0648799764624398
};

\addplot [semithick, red, mark=triangle*, mark size=3, mark options={solid}, only marks]
table {%
0.933333333333333 0.851505902157846
0.875 0.647918228262112
0.875 1.53307555912769
0.861111111111111 1.56905722565129
0.861111111111111 0.59101091418013
0.84375 0.659541733999048
0.839285714285714 0.929053372179315
0.833333333333333 0.791434239001155
0.830357142857143 0.708418317639754
0.821428571428571 0.39209414397191
0.8 0.610894526010398
0.790178571428571 0.700000993361912
0.787330316742081 0.874639002974763
0.776018099547511 0.461103344479193
0.772727272727273 1.05421566365227
0.75 0.610178850856362
0.75 0.938575010230195
0.739234449760765 0.992144749056266
};
\node at (axis cs:0.772727272727273, 1.05421566365227) [pin={80:${G_0^-}^*$}] {};

\addplot [semithick, red, mark=triangle*, mark options={rotate=180}, mark size=3, only marks]
table {%
0.763392857142857 0.183864346736873
};
\node at (axis cs:0.763392857142857,0.183864346736873) [pin={150:$G_1^-$}] {};

\addplot [semithick, red, mark=pentagon*, mark size=3, mark options={solid}, only marks]
table {%
0.758928571428571 0.404187215174276
};
\node at (axis cs:0.758928571428571, 0.404187215174276) [pin={150:$G_1^+$}] {};
\end{axis}

\end{tikzpicture}}
	\else
	\resizebox{0.8\columnwidth}{!}{\includegraphics[]{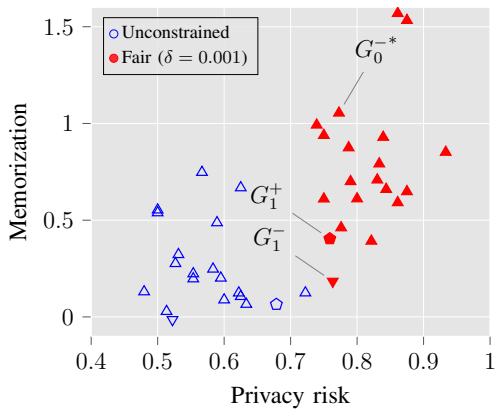}}
	\stepcounter{figureNumber}
	\fi
  \caption{\small The most vulnerable points on fair models (trained on synthetic data). We find the top 20 vulnerable points that have the highest privacy risk on fair models.  For each vulnerable point, we show its privacy risk and the memorization of models before (blue color) and after (red color) imposing fairness constraints. The marker shows which subgroup a point belongs to.}
  \label{fig:vulnerable_points}
\end{figure}

\subsection{Experimental Results on Synthetic Data}

\paragraphbe{Data and Models}
We generate synthetic datasets, of size $2,500$ records, similar to the prior work on analyzing fairness~\cite{zafar2019fairness}.  Specifically, we generate binary sensitive attributes, for each record, from a Bernoulli distribution ${P_g = \Pr[G=g]}$, for $g \in \{0, 1\}$.  We generate binary labels from a Bernoulli distributions ${P_g^y = \Pr[Y=y|G=g]}$, for $y \in \{-, +\}$ and for all $g \in \{0, 1\}$.  We generate a 2-dimensional feature vector from four different Gaussian distributions:
\begin{align}\label{eq:data_distribution}
	\text{For subgroup } G_{0}^-: ~&X \sim \N([0,-1],[7,1;1,7]) \nonumber \\
	\text{For subgroup } G_{1}^-: ~&X \sim \N([-5,0],[5,1;1,5]) \nonumber  \\
	\text{For subgroup } G_{0}^+: ~&X \sim \N([1,2],[5,2;2,5]) \nonumber  \\
	\text{For subgroup } G_{1}^+: ~&X \sim \N([2,3],[10,1;1,4])
\end{align}
where, $\N(\mu,\Sigma)$ represents a Gaussian distribution with mean vector $\mu$ and covariance matrix $\Sigma$.

We set $P_0=0.2$, $P_0^{-}=0.1$ and $P_1^{-}=0.5$. Accordingly, $P_1=0.8$, $P_0^+=0.9$ and $P_1^+=0.5$.  Group~$G_0$, thus, is the \emph{minority group} with a smaller number of samples.  The labels in this group are also unbalanced.  Subgroup~$G_0^-$ is the smallest subgroup.  

We use $50\%$ of the generated data for training and the rest for the testing.  We repeat this process (of splitting the data) $30$ times, and train $30$ unconstrained and fair models for each training set.  Each data point in our synthetic dataset, on average, appears in the training set of the model in $15$ experiments.  We report the average privacy cost over the $30$ experiments. 

We train fully connected neural network (NN) models with $3$ hidden layers with size~$\{32, 16, 8\}$, for both unconstrained and fair models. We use Adam Optimizer with learning rate~$0.001$.

\paragraphbe{Stronger inference attacks: adapting membership inference attacks to each group}
Membership inference attacks use a threshold on the loss of the model on a data point to infer whether the data point belongs to the training set.  It is however easier to distinguish members of the training set from non-members \emph{within} a subgroup.  Given that the adversary already has the information about the label and the group membership of a data point, he can take advantage of it to construct a stronger inference attack (compared with the prior work \cite{shokri2017membership,yeom2018privacy, sablayrolles2019white}).  Table~\ref{tab:multiple_threshold} shows the privacy risk for all subgroups when we use a single loss threshold (as in the prior work) versus using a separate attack (with a different loss threshold) for each subgroup. For each experiment, we report the average of $30$ runs.  The results show that using a separate attack threshold for each subgroup effectively increases the attack accuracy, which results in a better estimation of the privacy risk.


{
	\renewcommand{\arraystretch}{1.15}

\begin{table}[t!]
\caption{\small Accuracy and fairness gap of unconstrained and fair models (with three different fairness constraints $\delta$) on the synthetic datasets. The ``Train $\Delta$'' and ``Test $\Delta$'' columns show the fairness gap (as defined in~\eqref{eq:eo-fairness}) on the training and test data.} \label{tab:prediction_acc_syn}

\begin{tabularx}{\columnwidth}{
 c
>{\centering\arraybackslash}X c
>{\centering\arraybackslash}X c
>{\centering\arraybackslash}X c
>{\centering\arraybackslash}X c
}
\toprule
Model & Train acc& Test acc & Train $\Delta$ & Test $\Delta$ \\ \hline
Unconstrained& 86.2\%& 85.5\%& 0.373&0.430\\
Fair ($\delta=0.1$)& 89.1\%&87.8\%&0.105&0.332\\
Fair ($\delta=0.01$)&85.6\% &84.0\%&0.014 &0.283\\
Fair ($\delta=0.001$)&84.5\%&83.8\%&0.001&0.275\\
\bottomrule
\label{tab:accuracy_synthetic}
\end{tabularx}

\end{table}
}


{\renewcommand{\arraystretch}{1.15}
\begin{table}[t!]
\caption{\small Accuracy of membership inference attacks with a fixed loss threshold for all data points, versus using multiple thresholds one for each subgroup - Synthetic dataset.}
\begin{tabularx}{\columnwidth}{
c
c
>{\centering\arraybackslash}X c
>{\centering\arraybackslash}X c
>{\centering\arraybackslash}X c
>{\centering\arraybackslash}X c
}
\toprule
 Model & Attack & $G_{0}^-$ & $G_{1}^-$ & $G_{0}^+$ & $G_{1}^+$  \\ \hline
\multirow{2}{*}{Unconstrained}&
Single&52.9\%&51.2\%& 51.8\%& 51.2\% \\
&Group-based&61.8\% &52.8\%& 52.4\%& 52.2\% \\
\hline\hline
\multirow{2}{*}{Fair ($\delta=0.001$)}&
Single&60.8\% &51.9\%& 51.6\%& 50.8\% \\
&Group-based&69.2\%& 53.4\%&52.5\%& 51.6\% \\
\bottomrule
\end{tabularx}
\label{tab:multiple_threshold}
\end{table}
}

\paragraphbe{Privacy cost of fairness: unprivileged subgroups experience the largest cost}
Table~\ref{tab:accuracy_synthetic} shows the performance of unconstrained and fair models with different fairness constraints.  Table~\ref{tab:other-fair-notions} shows the prediction accuracy of unconstrained models for each subgroup.

We refer to $G_0^-$ as the \emph{unprivileged subgroup}, as it has the worst accuracy ($41.6\%$).  Compared with the samples from $G_1^-$ ($84.6\%$). We use asterisk on the unprivileged subgroup $G_1^{-*}$ to distinguish it from other subgroups.

Figure~\ref{fig:privacy_gap_four_subgroups} shows the histogram of individual privacy cost, across all subgroups.  The plots illustrate the imposed privacy risk due to fairness constraints, notably for the unprivileged subgroup.  We observe that $G_0^{-}$, unlike other subgroups, has a larger fraction of samples with positive privacy cost.  

To further study the privacy cost for individual data points, we identify $20$ vulnerable points with the largest privacy risk on fair models.  Figure~\ref{fig:vulnerable_points} shows the privacy risk for these data points before and after imposing fairness constraints.  We observe that these points are mainly from the unprivileged subgroup $G_0^{-}$.  In addition, the fairness constraints increase the privacy risk of these data points significantly. For example, the privacy risk of a data point increases from 0.63 to 0.93 after imposing fairness constraints.


\begin{figure}[t!]
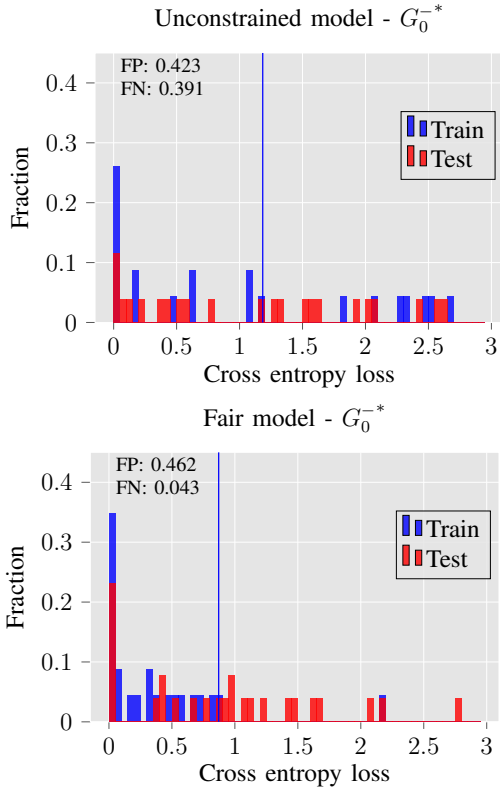

	\centering
	\if\compileFigures1
  \resizebox{0.8\columnwidth}{!}{\input{figure_scripts/fig_loss_distribution_unconstrained}}
  \resizebox{0.8\columnwidth}{!}{\input{figure_scripts/fig_loss_distribution_fair}}
	\else
	\resizebox{0.8\columnwidth}{!}{\includegraphics[]{fig/\filename-figure\thefigureNumber.pdf}}
	\stepcounter{figureNumber}
	\resizebox{0.8\columnwidth}{!}{\includegraphics[]{fig/\filename-figure\thefigureNumber.pdf}}
	\stepcounter{figureNumber}
	\fi
  \caption{\small Loss distribution of an unconstrained model and a fair model on subgroup $G_{0}^-$. The vertical blue line shows the loss threshold used in the membership inference attack.  FN is the false negative, and FP is the false positive rate for the attack.}
  \label{fig:loss-distribution-paper}
\end{figure}

Figure~\ref{fig:loss-distribution-paper} compares the loss distribution of fair and unconstrained models on their training and test data from subgroup~$G_{0}^-$.  We observe that \emph{members of training set are more distinguishable from non-members on fair models} compared with unconstrained models.  Thus, on this unprivileged subgroup, the adversary achieves a very low false-negative rate on fair model.


\begin{figure}[t!]
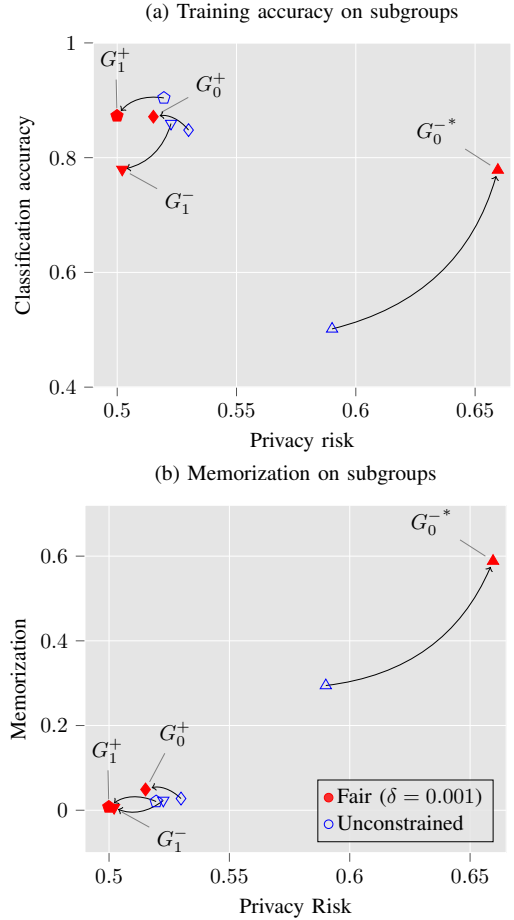

	\centering
	\if\compileFigures1
  \resizebox{0.8\columnwidth}{!}{
\begin{tikzpicture}
\begin{groupplot}[group style={group size=4 by 1}]
\nextgroupplot[
axis background/.style={fill=white!89.8039215686275!black},
axis line style={white},
tick align=outside,
tick pos=left,
title={(a) Training accuracy on subgroups},
x grid style={white},
xlabel={Privacy risk},
xmajorgrids,
xmin=0.49, xmax=0.665,
xtick style={color=white!33.3333333333333!black},
y grid style={white},
ylabel={Classification accuracy},
ymajorgrids,
ymin=0.4, ymax=1,
ytick style={color=white!33.3333333333333!black}
]

\addplot [red, mark=triangle*, mark size=3, only marks]
table {%
	0.659541929317076 0.778428120408528
};
\addplot [blue, mark=triangle, mark size=3, only marks]
table {%
	0.590058403525106 0.501509803954923
};
\node at (axis cs:0.659541929317076, 0.778428120408528) [pin={150:${G_0^-}^*$}] {};
\path[bend left, <-,shorten <=3pt] (axis cs:0.659541929317076, 0.778428120408528) edge (axis cs:0.590058403525106, 0.501509803954923);

\addplot [red, mark=diamond*, mark size=3, only marks]
table {%
	0.515169499363402 0.871360342142756
};
\addplot [blue, mark=diamond, mark size=3, only marks]
table {%
	0.5299053508436 0.848056137561798
};
\node at (axis cs:0.515169499363402, 0.871360342142756) [pin={20:$G_0^+$}] {};
\path[bend left, <-,shorten <=3pt] (axis cs:0.515169499363402, 0.871360342142756) edge (axis cs:0.5299053508436, 0.848056137561798);

\addplot [red, mark=triangle*, mark size=3, mark options={rotate=180}, only marks]
table {
	0.502107639579651 0.77947873556035
};
\addplot [blue, mark=triangle, mark size=3, mark options={rotate=180}, only marks]
table {
	0.522554930079399 0.859454666231162
};
\node at (axis cs:0.502107639579651, 0.77947873556035) [pin={340:$G_1^-$}] {};
\path[bend right, <-,shorten <=2pt] (axis cs:0.502107639579651, 0.77947873556035
) edge (axis cs:0.522554930079399, 0.859454666231162);

\addplot [red, mark=pentagon*, mark size=3, only marks]
table {
	0.499914838524796 0.87279801331898
};
\addplot [blue, mark=pentagon, mark size=3, only marks]
table {
	0.519570256222031 0.904062926769257
};
\node at (axis cs:0.499914838524796, 0.87279801331898) [pin={92:$G_1^+$}] {};

\path[bend left, <-,shorten <=3pt] (axis cs:0.499914838524796, 0.87279801331898) edge (axis cs:0.519570256222031, 0.904062926769257);

\end{groupplot}

\end{tikzpicture}}
  \resizebox{0.8\columnwidth}{!}{
\begin{tikzpicture}

\begin{axis}[
	axis background/.style={fill=white!89.8039215686275!black},
	axis line style={white},
	legend pos = south east,
	legend cell align={left},
	legend style={fill opacity=0.8, draw opacity=1, text opacity=1, fill=white!89.8039215686275!black},
	tick align=outside,
	tick pos=left,
	title={(b) Memorization on subgroups},
	x grid style={white},
	xlabel={Privacy Risk},
	xmajorgrids,
	xmin=0.49, xmax=0.665,
	xtick style={color=white!33.3333333333333!black},
	y grid style={white},
	ylabel={Memorization},
	ymajorgrids,
	ymin=-0.1, ymax=0.72,
	ytick style={color=white!33.3333333333333!black}
	]

	\addlegendimage{red, mark=*, mark size=2, only marks}
	\addlegendentry{Fair ($\delta=0.001$)};
	\addlegendimage{blue, mark=o, mark size=2, only marks}
	\addlegendentry{Unconstrained};

	\addplot [red, mark=triangle*, mark size=3, only marks]
	table {%
		0.659541929317076 0.58834295776175
	};
	\addplot [blue, mark=triangle, mark size=3, only marks]
	table {%
		0.590058403525106 0.294505586950585
	};
	\node at (axis cs:0.659541929317076, 0.58834295776175) [pin={150:${G_0^-}^*$}] {};
	\path[bend left, <-,shorten <=3pt] (axis cs:0.659541929317076, 0.58834295776175) edge (axis cs:0.590058403525106, 0.294505586950585);

	\addplot [red, mark=diamond*, mark size=3, only marks]
	table {%
		0.515169499363402 0.0489986647797789
	};
	\addplot [blue, mark=diamond, mark size=3, only marks]
	table {%
		0.5299053508436 0.0275266523301932
	};
	\node at (axis cs:0.515169499363402, 0.0489986647797789) [pin={80:$G_0^+$}] {};
	\path[bend left, <-,shorten <=3pt] (axis cs:0.515169499363402, 0.0489986647797789) edge (axis cs:0.5299053508436, 0.0275266523301932);

	\addplot [red, mark=triangle*, mark size=3, mark options={rotate=180}, only marks]
	table {
		0.502107639579651 0.00591825300450214
	};
	\addplot [blue, mark=triangle, mark size=3, mark options={rotate=180}, only marks]
	table {
		0.522554930079399 0.0231706561606779
	};
	\node at (axis cs:0.502107639579651, 0.00591825300450214) [pin={340:$G_1^-$}] {};
	\path[bend right, <-,shorten <=2pt] (axis cs:0.502107639579651, 0.00591825300450214) edge (axis cs:0.522554930079399, 0.0231706561606779);

	\addplot [red, mark=pentagon*, mark size=3, only marks]
	table {
		0.499914838524796 0.00776169668467611
	};
	\addplot [blue, mark=pentagon, mark size=3, only marks]
	table {
		0.519570256222031 0.02058045284326
	};
	\node at (axis cs:0.499914838524796, 0.00776169668467611) [pin={92:$G_1^+$}] {};

	\path[bend left, <-,shorten <=3pt] (axis cs:0.499914838524796, 0.00776169668467611) edge (axis cs:0.519570256222031, 0.02058045284326);

\end{axis}
\end{tikzpicture}}
	\else
	\resizebox{0.8\columnwidth}{!}{\includegraphics[]{fig/\filename-figure\thefigureNumber.pdf}}
	\stepcounter{figureNumber}
	\resizebox{0.8\columnwidth}{!}{\includegraphics[]{fig/\filename-figure\thefigureNumber.pdf}}
	\stepcounter{figureNumber}
	\fi
  \caption{\small Memorization and training accuracy of fair and unconstrained models on all subgroups - trained on synthetic data. The blue/red color show the results on the unconstrained/fair models. The markers represent different subgroups.}
  \label{fig:memorization}
\end{figure}

\paragraphbe{Training data memorization on fair models}
To further study why fairness constraints increase the privacy risk, we take a closer look at the memorization of individual training data by unconstrained versus fair models.

Figure~\ref{fig:memorization}(a) compares the average training accuracy of unconstrained and fair models for all subgroups over $30$ experiments.  The unconstrained models have a low accuracy on subgroup $G_{0}^-$ (which, in comparison with accuracy on $G_{1}^-$, shows the unfairness of the model according to the equalized odds measure).  After imposing the fairness constraints, the training accuracy on $G_{0}^-$ increases from $50.1\%$ to $78.8\%$.  We study if this improvement in accuracy is due to the training data memorization. 

We quantify the memorization of training data points, as the difference in the model's loss when the data point is in the training set versus the case where the data is not in the training set~\cite{feldman2020does}.

Figure~\ref{fig:memorization}(b) shows that the memorization of fair models on subgroup $G_{0}^-$ is $0.58$, which is two times larger than that of unconstrained models.  It shows that fair models memorize the points from $G_{0}^-$ instead of learning a general pattern about them.  On the contrary, fairness constraints only barely change the privacy risk and the memorization for other subgroups.  Accordingly, the privacy cost on other subgroups is low. Overall, these results show that fair constraints impose ``privacy unfairness''.


\begin{figure}[t!]
	\centering
	\if\compileFigures1
  \resizebox{0.8\columnwidth}{!}{
\begin{tikzpicture}
\large
\definecolor{color0}{rgb}{0.886274509803922,0.290196078431373,0.2}
\begin{axis}[
axis background/.style={fill=white!89.8039215686275!black},
axis line style={white},
tick align=outside,
tick pos=left,
title={\(\displaystyle {G_{0}^-}^*\)},
x grid style={white},
xlabel={Training accuracy gain},
xmajorgrids,
xmin=-0.2, xmax=1,
xtick style={color=white!33.3333333333333!black},
y grid style={white},
ylabel={Privacy cost (true positive rate)},
ymajorgrids,
ymin=-0.2, ymax=1,
ytick style={color=white!33.3333333333333!black}
]
\addplot [only marks, mark=*, draw=color0, fill=color0, colormap/viridis]
table{%
x                      y
0.482590907095531 0.769230769230769
0.510803132919645 0.466666666666667
0.475964562591994 0.111111111111111
-0.102402075959074 0.0909090909090909
0.20552169084549 0
-0.0207022659296667 0.05
0.315092817581641 0.15
0.402303746768406 0.357142857142857
0.0761688072234392 0.0625
0.0554111282030741 0
0.351993703537566 0.666666666666667
0.530699304504016 0.352941176470588
0.642378759560495 0.666666666666667
0.00492282211780548 0
0.133351365202352 0
0.753338810412738 0.733333333333333
0.526684359215202 0.625
-0.00123981498436487 0
0.642538247356166 0.529411764705882
-0.0252899229526523 0.0833333333333333
0.146683056246151 0.136363636363636
0.0272070401824234 0.2
0.855746491364444 0.823529411764706
0.82066907500422 0.625
0.139928212761879 0.1
0.441067183382601 0.5
0.0310560299290551 0
0.214496891945601 0.1875
0.0516108069568872 0.0625
0.638317012290847 0.6875
0.545286899785481 0.636363636363636
0.45524862408638 0.333333333333333
0.177300623673144 -0.133333333333333
0.0113475365298135 0
0.544916903261684 0.625
0.346458025276661 0.125
0.322828842326999 0.0625
0.236585895820083 0.0833333333333334
0.337195557445638 0.384615384615385
0.765536840145405 0.615384615384615
0.349571192209696 0.1875
0.0840162111497498 0
0.363221889734268 0.2
0.234202069119559 0.266666666666667
0.437765799748388 0.625
0.281544765457511 0.125
0.43291611772534 0.533333333333333
0.0515759338188065 0.125
0.409945728377209 0.555555555555556
};
\end{axis}

\end{tikzpicture}}
	\else
	\resizebox{0.8\columnwidth}{!}{\includegraphics[]{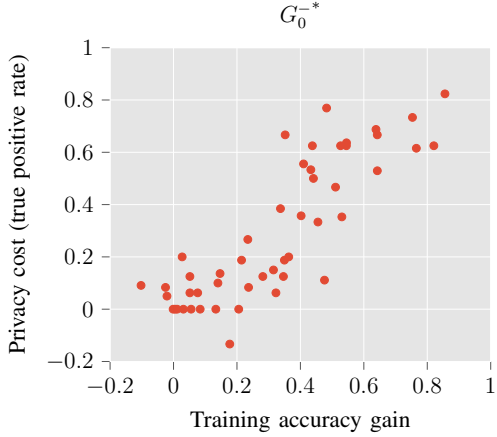}}
	\stepcounter{figureNumber}
	\fi
  \caption{\small  Accuracy gain versus privacy cost on the unprivileged subgroup $G_{0}^-$.  Each point in the plot represents a data point in the training dataset.  The training accuracy gain is the difference in the training accuracy between fair and unconstrained models.  The y-axis is the difference in the attacker's true positive rate between fair and unconstrained models on each training point.
  }
  \label{fig:acc_privacy_per_point}
\end{figure}

\paragraphbe{Accuracy-privacy trade-off on training data}
Figure~\ref{fig:acc_privacy_per_point} shows the effect of fairness constraints on the individual privacy cost and accuracy gain of training data in~$G_{0}^-$.

We measure the gain in training accuracy for each data point as the difference in prediction accuracy between fair models and unconstrained models on that data point. Thus, a positive accuracy gain implies that fairness constraints improve accuracy.  As we are analyzing the performance change on the training dataset, we only report the true positive rate of the adversary.  Recall that the true positive rate of the adversary reflects the probability of correctly predicting the membership of a data point when it \emph{is} a member of the training dataset.  The figure shows that there is a clear correlation between the accuracy gain and the privacy cost on training data in the unprivileged group.


\begin{figure}[t!]
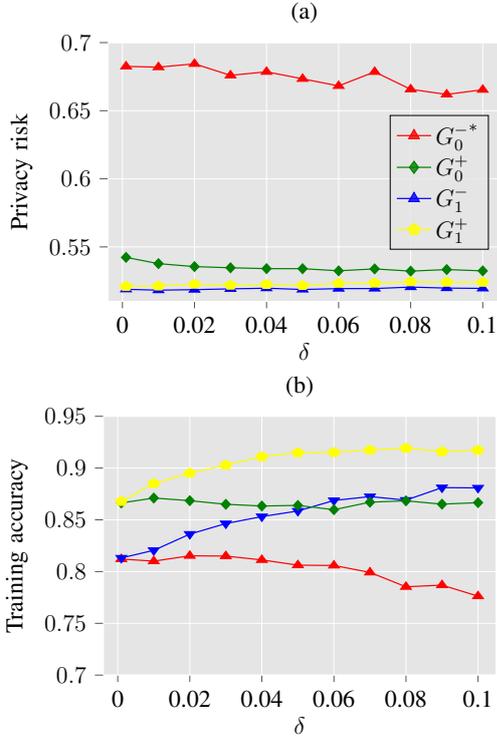

	\centering
	\if\compileFigures1
  \resizebox{0.8\columnwidth}{!}{
\begin{tikzpicture}
\large
\definecolor{color0}{rgb}{1,1,0}

\begin{axis}[
axis background/.style={fill=white!89.8039215686275!black},
axis line style={white},
legend cell align={left},
legend style={fill opacity=0.8, draw opacity=1, text opacity=1, at={(0.97,0.9)},  fill=white!89.8039215686275!black},
tick align=outside,
tick pos=left,
title={(a)},
x grid style={white},
xlabel={\(\displaystyle \delta\)},
xmajorgrids,
xmin=-0.00395, xmax=0.10495,
xtick style={color=white!33.3333333333333!black},
y grid style={white},
ylabel={Privacy risk},
yscale=.8,
title style={yshift=0.9cm},
xtick={0,0.02,0.04,0.06,0.08,0.1},
xticklabels={0,0.02,0.04,0.06,0.08,0.1},
ymajorgrids,
ymin=0.51, ymax=0.7,
ytick style={color=white!33.3333333333333!black}
]

\addplot [semithick, red, mark=triangle*, mark size=3, mark options={solid}]
table {%
	0.001 0.682527245916797
	0.01 0.681977077322348
	0.02 0.684410227658725
	0.03 0.67596322202113
	0.04 0.678562983638363
	0.05 0.673399769651565
	0.06 0.668220831087506
	0.07 0.678520659492365
	0.08 0.665722606466578
	0.09 0.661955049998573
	0.1 0.66540895265863
};
\addlegendentry{${G_0^-}^*$}

\addplot [semithick, green!50.1960784313725!black, mark=diamond*, mark size=3, mark options={solid}]
table {%
	0.001 0.542328344534026
	0.01 0.537730394508881
	0.02 0.535549439181551
	0.03 0.534699822130667
	0.04 0.534067007150617
	0.05 0.534035094290452
	0.06 0.532474500669468
	0.07 0.533871918977946
	0.08 0.532227270816103
	0.09 0.533384091634627
	0.1 0.532420269511157
};
\addlegendentry{$G_0^+$}

\addplot [semithick, blue, mark=triangle*, mark options={rotate=180}, mark size=3, mark options={solid}]
table {%
	0.001 0.518877983699575
	0.01 0.518252590114666
	0.02 0.518633587762617
	0.03 0.519206373817634
	0.04 0.519697893725577
	0.05 0.518739901964217
	0.06 0.519392188177733
	0.07 0.51944844728642
	0.08 0.520509565500657
	0.09 0.519783687392703
	0.1 0.519541816107936
};
\addlegendentry{$G_1^-$}

\addplot [semithick, color0, mark=pentagon*, mark size=3, mark options={solid}]
table {%
	0.001 0.521050595869252
	0.01 0.521316109864138
	0.02 0.522615869869885
	0.03 0.52192502641765
	0.04 0.522454481150084
	0.05 0.521665850859944
	0.06 0.523297276864054
	0.07 0.523437418928598
	0.08 0.52441395829417
	0.09 0.524101056451973
	0.1 0.524064055374808
};
\addlegendentry{$G_1^+$}
\end{axis}

\end{tikzpicture}}
    \resizebox{0.8\columnwidth}{!}{
\begin{tikzpicture}
\large
\definecolor{color0}{rgb}{1,1,0}
\begin{axis}[
axis background/.style={fill=white!89.8039215686275!black},
axis line style={white},
legend cell align={left},
legend style={fill opacity=0.8, draw opacity=1, text opacity=1, fill=white!89.8039215686275!black},
tick align=outside,
tick pos=left,
title={(b)},
x grid style={white},
xlabel={\(\displaystyle \delta\)},
legend pos = north west,
xmajorgrids,
xmin=-0.00395, xmax=0.10495,
xtick style={color=white!33.3333333333333!black},
xtick={0,0.02,0.04,0.06,0.08,0.1},
xticklabels={0,0.02,0.04,0.06,0.08,0.1},
y grid style={white},
ylabel={Training accuracy},
ymajorgrids,
yscale=0.8,
title style={yshift=0.9cm},
ymin=0.7, ymax=0.95,
ytick style={color=white!33.3333333333333!black}
]
\addplot [semithick, red, mark=triangle*, mark size=3, mark options={solid}]
table {%
	0.001 0.812101355562599
	0.01 0.810068636267085
	0.02 0.815261863060169
	0.03 0.814969285820686
	0.04 0.811273673799728
	0.05 0.806187598661506
	0.06 0.805893897667487
	0.07 0.799190935079821
	0.08 0.785290349916306
	0.09 0.786910472409413
	0.1 0.776268836097071
};
\addplot [semithick, blue, mark=triangle*, mark options={rotate=180}, mark size=3]
table {%
	0.001 0.81314785707328
	0.01 0.820533652496122
	0.02 0.836191896577109
	0.03 0.846364336170575
	0.04 0.853133740928161
	0.05 0.858512681915528
	0.06 0.868683998150193
	0.07 0.872446052404953
	0.08 0.869010483625805
	0.09 0.881095620915039
	0.1 0.880828937332345
};
\addplot [semithick, green!50.1960784313725!black, mark=diamond*, mark size=3, mark options={solid}]
table {%
	0.001 0.866298422570041
	0.01 0.871055379262556
	0.02 0.868484886298075
	0.03 0.865007836559115
	0.04 0.863326761955459
	0.05 0.863917572351103
	0.06 0.859764099762525
	0.07 0.867042073279311
	0.08 0.868358217942145
	0.09 0.865128863900602
	0.1 0.866591513300079
};
\addplot [semithick, color0, mark=pentagon*, mark size=3, mark options={solid}]
table {%
	0.001 0.867738475745045
	0.01 0.885073228821788
	0.02 0.895113068124093
	0.03 0.902821624100192
	0.04 0.910954364164098
	0.05 0.914783219899484
	0.06 0.91513132436277
	0.07 0.917390388667263
	0.08 0.919152697819749
	0.09 0.915895896086773
	0.1 0.917287151181103
};
\end{axis}

\end{tikzpicture}}
	\else
	\resizebox{0.8\columnwidth}{!}{\includegraphics[]{fig/\filename-figure\thefigureNumber.pdf}}
	\stepcounter{figureNumber}
		\resizebox{0.8\columnwidth}{!}{\includegraphics[]{fig/\filename-figure\thefigureNumber.pdf}}
	\stepcounter{figureNumber}
	\fi
  \caption{\small (a) The effect of the enforced fairness level $\delta$ on the privacy risk of fair models for different subgroups. (b) The effect of enforced fairness level on the classification accuracy of fair models for different subgroups.}
  \label{fig:trade-offs-privacy-accuracy}
\end{figure}


\begin{figure}[t!]
	\centering
	\if\compileFigures1
  \resizebox{0.8\columnwidth}{!}{
\begin{tikzpicture}

\begin{axis}[
axis background/.style={fill=white!89.8039215686275!black},
axis line style={white},
legend cell align={left},
legend style={fill opacity=0.8, draw opacity=1, text opacity=1, at={(0.03,0.97)}, anchor=north west, fill=white!89.8039215686275!black},
tick align=outside,
tick pos=left,
title={Memorization and privacy risk on \(\displaystyle {G_{0}^-}^*\)},
x grid style={white},
xlabel={Privacy Risk},
xmajorgrids,
xmin=0.55, xmax=0.68,
xtick style={color=white!33.3333333333333!black},
y grid style={white},
ylabel={Memorization},
ymajorgrids,
ymin=0.2, ymax=0.6,
ytick style={color=white!33.3333333333333!black}
]
\addplot [semithick, blue, mark=o, mark size=3, mark options={solid}, only marks]
table {%
0.590058403525106 0.294505586950585
};
\addlegendentry{Unconstrained}
\addplot [semithick, red, mark=x, mark size=4, mark options={solid}, only marks]
table {%
0.632605348604906 0.509735921187993
};
\addlegendentry{Fair ($\delta=0.1$)}
\addplot [semithick, red, mark=asterisk, mark size=4, only marks]
table {%
0.655743223862972 0.557107030392323
};
\addlegendentry{Fair ($\delta=0.01$)}
\addplot [semithick, red, mark=triangle*, mark size=4, only marks]
table {%
0.659541929317076 0.58834295776175
};
\addlegendentry{Fair ($\delta=0.001$)}
\end{axis}

\end{tikzpicture}}
	\else
	\resizebox{0.8\columnwidth}{!}{\includegraphics[]{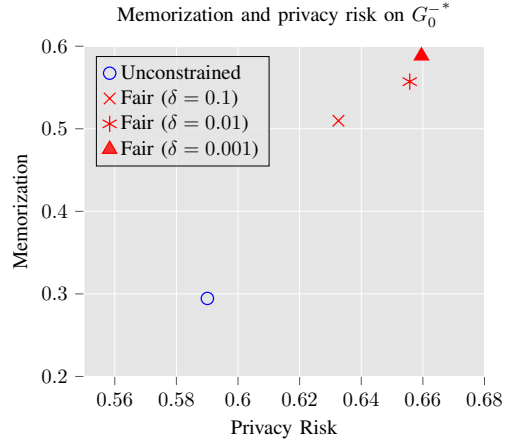}}
	\stepcounter{figureNumber}
	\fi
  \caption{\small The effect of the enforced fairness gap $\delta$ on the privacy risk and memorization of the model - Synthetic dataset.}
  \label{fig:enforced_fairness_gap}
\end{figure}


\begin{figure}[t!]
	\centering
	\if\compileFigures1
  \resizebox{0.8\columnwidth}{!}{
\begin{tikzpicture}
\large
\begin{axis}[
axis background/.style={fill=white!89.8039215686275!black},
axis line style={white},
legend cell align={left},
legend style={fill opacity=0.8, draw opacity=1, text opacity=1, at={(0.03,0.97)}, anchor=north west, draw=white!80!black, fill=white!89.8039215686275!black},
ytick={-0.05, 0, 0.05, 0.1, 0.15, 0.2},
yticklabels={-0.05, 0, 0.05, 0.1, 0.15, 0.2},
tick align=outside,
tick pos=left,
yscale=.8,
x grid style={white},
xlabel={ \(\displaystyle \Delta(S,M_s)\)},
x label style={at={(axis description cs:0.5,-0.15)},anchor=north},
y label style={at={(axis description cs:-0.15,.5)},anchor=south},
xmajorgrids,
xmin=0, xmax=0.6,
title style={yshift=0.9cm},
xtick style={color=white!33.3333333333333!black},
y grid style={white},
ylabel={Privacy cost on \(\displaystyle {G_{0}^-}^*\)},
ymajorgrids,
ymin=-0.05, ymax=0.2,
ytick style={color=white!33.3333333333333!black},
title={},
]
\addplot [semithick, red, mark=*, mark size=3, mark options={solid}, only marks]
table {%
0.553874389575329 0.136477550798659
};
\addplot [semithick, red, mark=*, mark size=3, mark options={solid}, only marks]
table {%
0.563580192518761 0.152031816892266
};
\addplot [semithick, red, mark=*, mark size=3, mark options={solid}, only marks]
table {%
0.467239689999702 0.112472338243026
};
\addplot [semithick, red, mark=*, mark size=3, mark options={solid}, only marks]
table {%
0.376826221226737 0.0490076557902646
};
\addplot [semithick, red, mark=*, mark size=3, mark options={solid}, only marks]
table {%
0.294108982173815 0.0279965163835991
};
\addplot [semithick, red, mark=*, mark size=3, mark options={solid}, only marks]
table {%
0.192857609204439 0.00557576617396727
};
\addplot [semithick, red, mark=*, mark size=3, mark options={solid}, only marks]
table {%
0.109598601007543 0.00411078464666992
};
\end{axis}

\end{tikzpicture}}
	\else
	\resizebox{0.8\columnwidth}{!}{\includegraphics[]{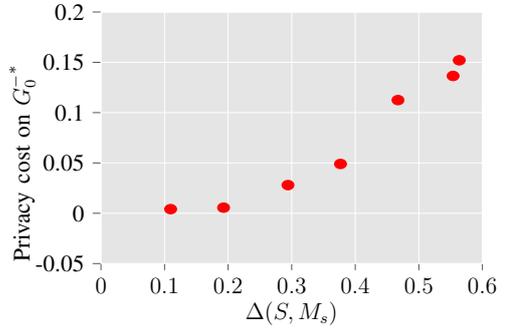}}
	\stepcounter{figureNumber}
	\fi
	\vspace{1em}
  \caption{\small The effect of the unconstrained model's fairness gap (which captures the unfairness that needs to be removed by the fair algorithm) on the privacy cost of $G_0^-$. The x-axis is the fairness gap of unconstrained models on the training dataset. }
  \label{fig:trade-offs}
\end{figure}

Figure~\ref{fig:trade-offs-privacy-accuracy} shows the effect of the enforced fairness gap~$\delta$ on the privacy risk and training accuracy of subgroups.  Recall that smaller $\delta$ makes the models less discriminatory on the training dataset.  We observe that, as $\delta$ decreases, accuracy as well as privacy risk for data points in $G_0^-$ increases.

We also analyze the effect of $\delta$ on memorization.  Figure~\ref{fig:enforced_fairness_gap} shows that, when a fair model is less discriminatory on its training data (i.e., $\delta$ is smaller), the memorization of data points in subgroup $G_{0}^-$ is larger (thus, the privacy risk is larger). 

\paragraphbe{Effect of underlying unfairness on privacy cost}
We generate multiple synthetic datasets, by varying the mean of distribution~\eqref{eq:data_distribution} for data in subgroup $G_{0}^-$, to control the underlying unfairness that this change causes on the unconstrained model.  Varying mean results in different separability between positive and negative samples, which influences the difficulty of learning an accurate model that distinguishes between $G_{0}^-$ and $G_{0}^+$ samples.  This influences the fairness gap on the unconstrained models.  In this setting, a large fairness gap means that subgroup $G_{0}^-$ has a worse accuracy compared with subgroup $G_{1}^-$.  

Figure~\ref{fig:trade-offs} shows the correlation between the fairness gap $\Delta(S,\M_\train)$ (in the unconstrained model) and the subgroup privacy cost for $G_0^-$.  We observe that a large fairness gap in the underlying unconstrained model results in a large privacy cost (of imposing fairness constraints).  In other words, when the unconstrained model is more discriminatory, and there is more need for a fairness mechanism, \emph{reducing accuracy disparity results in a significant privacy disparity for unprivileged groups}. 


\begin{figure*}[t!]
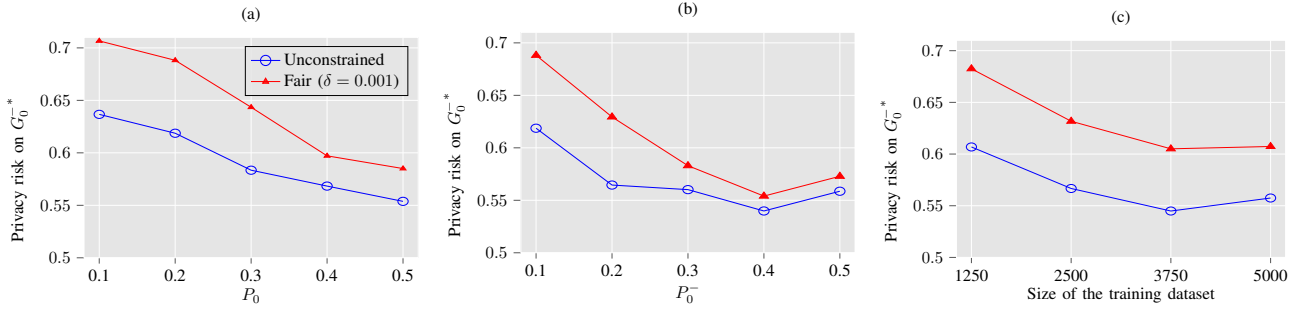

	\centering
	\if\compileFigures1
  \resizebox{0.32\textwidth}{!}{
\begin{tikzpicture}
\begin{axis}[
axis background/.style={fill=white!89.8039215686275!black},
axis line style={white},
legend cell align={left},
legend style={fill opacity=0.8, draw opacity=1, text opacity=1,  at={(0.97,1.2)}, fill=white!89.8039215686275!black},
tick align=outside,
tick pos=left,
title={(a)},
x grid style={white},
xlabel={\(\displaystyle P_0\)},
xmajorgrids,
yscale=.8,
xmin=-0.2, xmax=4.2,
xtick style={color=white!33.3333333333333!black},
xtick={0,1,2,3,4},
xticklabels={0.1,0.2,0.3,0.4,0.5},
y label style={at={(axis description cs:-0.15,.5)},anchor=south},
y grid style={white},
ylabel={Privacy risk on ${G_0^-}^*$},
ymajorgrids,
ymin=0.5, ymax=0.71,
y label style={at={(axis description cs:-0.15,.5)},anchor=south},
ytick style={color=white!33.3333333333333!black},
title style={yshift=0.9cm}
]
\addplot [semithick, blue, mark=o, mark size=3, mark options={solid}]
table {%
0 0.636666181041181
1 0.618678258824158
2 0.583458485226935
3 0.568322097725021
4 0.553772056358016
};
\addlegendentry{Unconstrained}
\addplot [semithick, red, mark=triangle*, mark size=2, mark options={solid}]
table {%
0 0.706643634143634
1 0.688171445533164
2 0.643442762495251
3 0.597142319830346
4 0.585021321001899
};
\addlegendentry{Fair ($\delta=0.001$)}
\end{axis}

\end{tikzpicture}}
  \resizebox{0.32\textwidth}{!}{
\begin{tikzpicture}
\begin{axis}[
axis background/.style={fill=white!89.8039215686275!black},
axis line style={white},
legend cell align={left},
legend style={fill opacity=0.8, draw opacity=1, text opacity=1, draw=white!80!black, fill=white!89.8039215686275!black},
tick align=outside,
tick pos=left,
title={(b)},
title style={yshift=0.9cm},
x grid style={white},
xlabel={\(\displaystyle P_0^-\)},
xmajorgrids,
xmin=-0.2, xmax=4.2,
xtick style={color=white!33.3333333333333!black},
xtick={0,1,2,3,4},
yscale=.8,
xticklabels={0.1,0.2,0.3,0.4,0.5},
y grid style={white},
ylabel={Privacy risk on ${G_0^-}^*$},
ymajorgrids,
ymin=0.5, ymax=0.71,
y label style={at={(axis description cs:-0.15,.5)},anchor=south},
ytick style={color=white!33.3333333333333!black}
]
\addplot [semithick, blue, mark=o, mark size=3]
table {%
0 0.618678258824158
1 0.56452980166117
2 0.560167626113528
3 0.539828023956387
4 0.558642724759473
};
\addplot [semithick, red, mark=triangle*, mark size=3, mark options={solid}]
table {%
0 0.688171445533164
1 0.629414924741535
2 0.582981740233789
3 0.554013539566394
4 0.572896233565345
};
\end{axis}

\end{tikzpicture}}
  \resizebox{0.32\textwidth}{!}{
\begin{tikzpicture}
\begin{axis}[
axis background/.style={fill=white!89.8039215686275!black},
axis line style={white},
legend cell align={right},
legend style={fill opacity=0.8, draw opacity=1, text opacity=1, at={(0.97,0.97)}, anchor=north east,  fill=white!89.8039215686275!black},
tick align=outside,
tick pos=left,
title={(c)},
title style={yshift=0.9cm},
x grid style={white},
xlabel={Size of the training dataset},
xmajorgrids,
xmin=-0.15, xmax=3.15,
xtick style={color=white!33.3333333333333!black},
yscale=.8,
xtick={0,1,2,3},
xticklabels={1250,2500,3750,5000},
y grid style={white},
ylabel={Privacy risk on ${G_0^-}^*$},
ymajorgrids,
ymin=0.5, ymax=0.71,
y label style={at={(axis description cs:-0.15,.5)},anchor=south},
ytick style={color=white!33.3333333333333!black}
]
\addplot [semithick, blue, mark=o, mark size=3, mark options={solid}]
table {%
0 0.606868489597071
1 0.566684072871136
2 0.544890041269145
3 0.557470853801268
};
\addplot [semithick, red, mark=triangle*, mark size=3, mark options={solid}]
table {%
0 0.682527245916797
1 0.631761769231358
2 0.605059271167149
3 0.607331221271576
};
\end{axis}

\end{tikzpicture}}
	\else
	\resizebox{0.32\textwidth}{!}{\includegraphics[height=10cm]{fig/\filename-figure\thefigureNumber.pdf}}
	\stepcounter{figureNumber}
		\resizebox{0.32\textwidth}{!}{\includegraphics[height=10cm]{fig/\filename-figure\thefigureNumber.pdf}}
	\stepcounter{figureNumber}
	\resizebox{0.32\textwidth}{!}{\includegraphics[height=10cm]{fig/\filename-figure\thefigureNumber.pdf}}
	\stepcounter{figureNumber}
	\fi
  \caption{\small Privacy risk on the unprivileged subgroup~$G_0^-$, for various faction of data in group~$G_0$, various fraction of data in subgroup~$G_0^-$, and for various training set size.}
  \label{fig:sample_size}
\end{figure*}

\paragraphbe{Effect of dataset size on privacy cost}
We generate synthetic data by varying the fraction of samples in $G_0$ (by controlling ${P_0=\Pr[G=0]}$) and the underprivileged subgroup $G_0^-$ (by controlling ${P_0^-=\Pr[y=-|G=0]}$).  Figure~\ref{fig:sample_size} shows the effect of the size of groups, and the size of the training set, on the privacy risk on the unprivileged subgroup.  As we expect, we observe that a smaller number of samples in the unprivileged subgroup~$G_0^-$ or its group~$G_0$ results in a higher privacy cost for the unprivileged group.  When the unprivileged subgroup~$G_0^-$ is relatively small with respect to subgroup~$G_0^+$, or the whole group~$G_0$ is small, the underlying fairness gap in an unconstrained model increases.  This leads to further memorization of the unprivileged group, thus higher privacy risk, when fairness constraints are enforced.  Figure~\ref{fig:sample_size}(c) shows that the training set size has a similar effect on both unconstrained and fair models, as long as the fraction of data in each subgroup remains the same.  The results show that, as expected, the privacy risk (with respect to~$G_0^-$) decreases as we increase the size of the training set.


{\renewcommand{\arraystretch}{1.15}
\begin{table}[t!]
\caption{\small Prediction accuracy and privacy risk for unconstrained models, and fair models trained using reduction approach~\cite{agarwal2018reductions} and post-processing (PP) approach~\cite{hardt2016equality}.}
\label{tab:other-fair-alg}
\begin{tabularx}{\columnwidth}{
cc
>{\centering\arraybackslash}X c
>{\centering\arraybackslash}X c
>{\centering\arraybackslash}X c
>{\centering\arraybackslash}X c
}
\toprule
Model & &${G_{0}^-}^*$ & $G_{1}^-$ & $G_{0}^+$ &$G_{1}^+$  \\
\bottomrule
\multirow{3}{*}{Unconstrained}& Train acc & 52.2\%&85.6\%&85.3\%&89.8\%\\
& Test acc & 39.1\%&84.7\%&86.2\%&89.5\%\\ 
& Privacy risk & 0.639&0.521&0.518&0.523\\

\hline\hline
\multirow{1}{*}{Fair (reduction) }
& Train acc & 80.1\%&80.2\%&88.4\%&88.5\% \\
\multirow{2}{*}{($\delta =0.001$)}& Test acc & 49.2\%&79.2\%&88.1\%&88.3\%\\
& Privacy risk & 0.688&0.521&0.542&0.518\\

\hline\hline
\multirow{1}{*}{Fair (PP) }& Train acc & 48.4\%&48.4\%&90.7\%&90.7\%\\
\multirow{2}{*}{($\delta =0$)}& Test acc & 41\% &47.9\% &89.4\% &90.5\% \\ 
& Privacy risk & 0.55 &0.507 &0.509 &0.504\\

\toprule
\end{tabularx}
\end{table}
}

\paragraphbe{Effect of fair algorithms}
Table~\ref{tab:other-fair-alg} compares the prediction accuracy and privacy risk of fair models using an in-processing approach~\cite{agarwal2018reductions} (which we analyze in all our experiments) and a post-processing approach~\cite{hardt2016equality}.  The post-processing algorithm does not improve the training accuracy of the unprivileged subgroup~$G_0^-$. Instead, it decreases the accuracy for another subgroup~$G_1^-$ from $85.6\%$ to $48.4\%$ to satisfy fairness constraints.  This means that, in the case of using post-processing approach, the fair model does not fit the unprivileged subgroup better than the unconstrained models. Thus, due to underfitting, the privacy risk of fair models is less than that of unconstrained models. 


{\renewcommand{\arraystretch}{1.15}
\begin{table}[t!]
\caption{\small Prediction accuracy and privacy risk for unconstrained models, versus fair models under different notions of fairness. We use the reduction approach~\cite{agarwal2018reductions} to train fair models and set $\delta=0.001$.}
\label{tab:other-fair-notions}
\begin{tabularx}{\columnwidth}{
cc
>{\centering\arraybackslash}X c
>{\centering\arraybackslash}X c
>{\centering\arraybackslash}X c
>{\centering\arraybackslash}X c
}
\toprule
Model & &${G_{0}^-}^*$ & $G_{1}^-$ & $G_{0}^+$ &$G_{1}^+$  \\
\bottomrule
\multirow{3}{*}{Unconstrained}
& Train acc & 47.8\%&85.1\%&85.8\%&89.2\% \\
& Test acc & 41.6\%&84.6\%&85.1\%&89\% \\ 
& Privacy risk & 0.607&0.521&0.529&0.522 \\

\hline\hline
\multirow{1}{*}{Fair}
& Train acc & 81.2\%&81.3\%&86.6\%&86.8\%\\
\multirow{2}{*}{(EO)}
& Test acc & 53.8\%&80.9\%&83.8\%&86.4\% \\
& Privacy risk & 0.683&0.519&0.542&0.521 \\

\hline\hline
\multirow{1}{*}{Fair}
& Train acc & 47.4\%&91\%&91.6\%&91.7\% \\
\multirow{2}{*}{(EOPP)}
& Test acc &39.1\%&90.1\%&89.7\%&91.3\% \\
& Privacy risk & 0.605&0.52&0.534&0.522 \\
\hline\hline

\multirow{1}{*}{Fair}
& Train acc & 83.1\%&83.2\%&85.5\%&91.8\% \\
\multirow{2}{*}{(FPP)}
& Test acc & 54.5\%&82.9\%&84\%&90.9\% \\
& Privacy risk & 0.679&0.518&0.535&0.523 \\

\toprule
\end{tabularx}
\end{table}
}

\paragraphbe{Effect of fairness notions}
We evaluate the privacy risk of fairness for two other notions of fairness: equal opportunity (EOPP), and false-positive parity (FPP). 

\begin{definition}[$\delta$ - Equal Opportunity Fairness~\cite{hardt2016equality}] \label{def:eopp_fairness}
	A classifier $\M$ satisfies $\delta$-Equal Opportunity condition with respect to the protected attribute $\G$, if for all $g, g' \in \G$, the false negative rate of the classifier in the group $\{G = g\}$ and $\{G = g' \}$ are within $\delta$ range of one another:
	\begin{align} \label{eq:eopp-fairness}
	\Delta(\M, \D) \triangleq  \nonumber \\
	\max_{\substack{y=+ \\ g,g' \in \G}} \bigg | &\Pr_\D[\M(X)\neq y|G=g, Y=y] \nonumber \\
	- &\Pr_\D[\M(X)\neq y|G=g', Y=y] \bigg | \le \delta .
	\end{align}
\end{definition}

By setting $y=-$ instead of $y=+$ in Equation~\eqref{eq:eopp-fairness}, we get the definition of false-positive parity.

In Table~\ref{tab:other-fair-notions}, we compare the prediction accuracy and privacy risk of unconstrained and fair models that satisfy different fairness notions.  Overall, we observe similar patterns across different group-fairness metrics.



{\renewcommand{\arraystretch}{1.15}
\begin{table}[t!]
\caption{\small The data partitioning based on the protected attributes, and the percentage of data in different subgroups for the real-world datasets.}
\begin{tabularx}{1\columnwidth}{
c
>{\centering\arraybackslash}X c
>{\centering\arraybackslash}X c
>{\centering\arraybackslash}X c
>{\centering\arraybackslash}X c
>{\centering\arraybackslash}X c
>{\centering\arraybackslash}X c
}
\toprule
 Name &  $G_{0}^-$& $G_{1}^-$ & $G_{0}^+$ &$G_{1}^+$  \\ \hline Bank (age)&2.2\%&85.2\%&0.7\%&12.0\%\\
COMPAS (race)&28.3\%&24.4\%&31.7\%&15.6\%\\
COMPAS (gender)&12.8\%&39.9\%&7.3\%&40.0\%\\
Law (race)&2.3\%&2.7\%&13.5\%&81.5\%\\
Law (gender)&2.5\%&2.5\%&41.4\%&53.6\%\\
\bottomrule
\end{tabularx}
\label{tab:real_data_statistic}
\end{table}
}


{\renewcommand{\arraystretch}{1.15}
\begin{table}[t!]
\caption{\small Accuracy and fairness gap $\Delta$ of unconstrained models and fair models (with different enforced fairness level~$\delta$) on the training and test dataset -- Decision tree model with max depth $10$.}
\label{tab:prediction_acc_real}
\begin{tabularx}{\columnwidth}{
c
c
 >{\centering\arraybackslash}X c
 >{\centering\arraybackslash}X c
 >{\centering\arraybackslash}X c
 >{\centering\arraybackslash}X c
 }
 \toprule
 Dataset & Model & Train acc & Test acc & Train $\Delta$ & Test $\Delta$  \\ \hline
\multirow{2}{*}{Bank}&
Unconstrained&92.5\%&87.8\%&0.063&0.074\\
&Fair ($\delta=0.1$)& 94.7\%&89.3\%&0.057&0.078\\
\multirow{1}{*}{(age)}&Fair ($\delta=0.01$)&94.6\%&89.3\%&0.011&0.063\\
&Fair ($\delta=0.001$)&94.6\%&89.3\%&0.001&0.066\\

 \hline\hline
\multirow{2}{*}{COMPAS}&
Unconstrained&72.4\%&60.1\%&0.097&0.133\\
&Fair ($\delta=0.1$)& 79.4\%&64.3\%&0.108&0.131\\
\multirow{1}{*}{(race)}&Fair ($\delta=0.01$)& 79\%&64\%&0.017&0.073\\
&Fair ($\delta=0.001$)&78.7\%&64\%&0.002&0.067\\

 \hline\hline

\multirow{2}{*}{COMPAS}&
Unconstrained&72.4\%&60.2\%&0.117&0.107\\
&Fair ($\delta=0.1$)&79.3\%&64.4\%&0.081&0.1 \\
\multirow{1}{*}{(gender)}&Fair ($\delta=0.01$)& 78.6\%&64.4\%&0.013&0.083\\
&Fair ($\delta=0.001$)&78.5\%&64.3\%&0.001&0.08\\

 \hline\hline
\multirow{2}{*}{Law}&
Unconstrained&95.9\%&92.2\%&0.236&0.165\\
&Fair ($\delta=0.1$)& 97.6\%&93.6\%&0.148&0.156\\
\multirow{1}{*}{(race)}&Fair ($\delta=0.01$)&97.5\%&93.4\%&0.018&0.12 \\
&Fair ($\delta=0.001$)&97.5\%&93.5\%&0.002&0.112\\

 \hline\hline
\multirow{2}{*}{Law}&
Unconstrained&95.9\%&92.2\%&0.035&0.039\\
&Fair ($\delta=0.1$)& 97.6\%&93.6\%&0.097&0.039\\
\multirow{1}{*}{(gender)}&Fair ($\delta=0.01$)& 97.6\%&93.6\%&0.019&0.04\\
&Fair ($\delta=0.001$)&97.6\%&93.6\%&0.002&0.033\\
 \bottomrule
 \end{tabularx}
 \label{tab:accuracy_real_data}
 \end{table}
 }

\subsection{Experimental Results on Real Data} \label{subsec:real-data}

\paragraphbe{Data and models}
We conduct experiments on the Law School dataset (Law)~\cite{lawdataset}\footnote{Downloaded from https://github.com/jjgold012/lab-project-fairness (Bechavod and Ligett, 2017)} (with $19$ features and $16,672$ data points), Bank Marketing dataset (Bank)~\cite{ucidataset} (with $58$ features and $24,391$ data points), and COMPAS dataset~\cite{compasdataset} (with $11$ features and $4,302$ data points).  We use the same preprocessing on these datasets as in IBM's AI Fairness 360~\cite{bellamy2018ai}.  For COMPAS and Law datasets, we consider two versions for each dataset, one where the protected attribute is ``race'' (white versus non-white) and the other where the protected attribute is ``gender'' (male versus female). For Bank dataset, the protected attribute is ``age'' (age $\geq 25$ versus age $ < 25$).  Table~\ref{tab:real_data_statistic} shows the distribution of data points across different subgroups. For all datasets, we use $50\%$ of the available data for training and the remaining $50\%$ for test. 

We train decision tree (DT) models, which are commonly evaluated in the fairness literature.  We train fair models using the reductions approach~\cite{agarwal2018reductions}.  For each experiment, we report the average results over $20$ runs.  Table~\ref{tab:accuracy_real_data} shows the performance of the unconstrained and fair models.  We defer readers to Table~\ref{tab:accuracy_subgroups_real_data} in Appendix~\ref{appendix-additional-results} for the detailed results for all subgroups.


{\renewcommand{\arraystretch}{1.15}
\begin{table}[t!]
\caption{\small Privacy risk of unconstrained and fair models (with $\delta = 0.001$) across different subgroups -- Decision tree models with max depth $10$.  We indicate the protected attribute for each dataset. Unprivileged subgroups are identified by asterisks. }
\label{tab:DT-privacy-cost}
\begin{tabularx}{\columnwidth}{
>{\centering\arraybackslash}X c
>{\centering\arraybackslash}X c
>{\centering\arraybackslash}X c
>{\centering\arraybackslash}X c
>{\centering\arraybackslash}X c
>{\centering\arraybackslash}X c
}
\toprule
Dataset & Model & $G_{0}^-$ & $G_{1}^-$ & $G_{0}^+$ &$G_{1}^+$  \\ \hline
Bank&
Unconstrained&$0.545$&0.516&0.645&${0.611}^*$\\
(age)&Fair &0.574&0.521&0.707&0.644\\

\hline\hline
COMPAS &
Unconstrained&$0.582$&0.565&0.579&${0.611}^*$\\
(race)&Fair&0.599&0.589&0.601&0.648\\
\hline\hline

COMPAS&
Unconstrained&0.583&${0.569}^*$&0.643&$0.576$\\
(gender)&Fair&0.572&0.591&0.643&0.599\\

\hline\hline
Law&
Unconstrained&0.726&${0.711}^*$&0.541&0.510\\
(race)&Fair&0.745&0.818&0.555&0.515\\
\hline\hline
Law&
Unconstrained&0.721&${0.724}^*$&0.514&0.514\\
(gender)&Fair&0.774&0.788&0.521&0.519\\
\bottomrule
\end{tabularx}
\end{table}
}

\paragraphbe{Privacy cost of fairness}
Table~\ref{tab:DT-privacy-cost} compares the privacy risk on fair and unconstrained models for different subgroups.  After imposing the fairness constraint, the gap between the privacy risk across different subgroups widens.  For instance, in the experiment on the Bank dataset, the difference between the privacy risk across subgroups~$G_{1}^-$ and~$G_{0}^+$ increases from $0.131$ (on the unconstrained model) to $0.186$ (on the fair model).  This shows the privacy risk disparity due to fairness, in the real data, similar to what we observe on synthetic data. 

One interesting observation is that the privacy cost on the COMPAS (gender) dataset is relatively smaller than that of the COMPAS (race) dataset.  Note that the two datasets are exactly the same.  The difference is that the fairness gap of the unconstrained model with respect to gender is $0.095$, which is much smaller than the fairness gap of the same model with respect to race.  Thus, the fairness constraint has less impact on the model that is trained on COMPAS (gender).  This results in less memorization, hence smaller privacy cost. 

On the Law (gender) dataset, we observe a different phenomenon.  Even though the fairness gap of the unconstrained model is small ($0.031$), we observe a high subgroup privacy cost ($0.064$) on  $G_1^-$.  We think the reason might be that the fair model is unable to learn generalizable patterns on this subgroup due to its relatively small size (as shown in Table~\ref{tab:real_data_statistic}); hence, it memorizes the data.  Table~\ref{tab:law-table} shows the prediction performance and subgroup privacy risk for the same dataset with a different protected attribute (Law (race) dataset).  By increasing the enforced level, we can increase the accuracy of the unprivileged subgroup, yet at the cost of its privacy. 


{\renewcommand{\arraystretch}{1.15}
\begin{table}
\caption{\small Prediction accuracy and privacy risk of unconstrained and fair models with different enforced fairness gap $\delta$ on decision tree models with max depth $10$ -- Law (race) dataset. }
\label{tab:law-table}
\begin{tabularx}{\columnwidth}{
c
c
>{\centering\arraybackslash}X c
>{\centering\arraybackslash}X c
>{\centering\arraybackslash}X c
>{\centering\arraybackslash}X c
}
\toprule
&Model & $G_{0}^-$& ${G_{1}^-}^*$ & $G_{0}^+$ &$G_{1}^+$  \\ \hline
\multirow{4}{*}{Train acc} &Unconstrained&70.1\%&43.4\%&99.0\%&99.8\%\\
&Fair ($\delta=0.1$)&64.7\%&49.7\%&99.3\%&99.8\%\\
&Fair ($\delta=0.01$)&55.6\%&53.8\%&99.6\%&99.8\%\\
&Fair ($\delta=0.001$)&54.5\%&54.3\%&99.6\%&99.7\%\\
\hline\hline
\multirow{4}{*}{Test acc} &Unconstrained&28.6\%&10.8\%&92.2\%&98.6\%\\
&Fair ($\delta=0.1$)&26.8\%&10.9\%&92.8\%&98.4\%\\
&Fair ($\delta=0.01$)&23.7\%&10.9\%&93.5\%&98.2\%\\
&Fair ($\delta=0.001$)&23.3\%&11.8\%&94.0\%&98.1\%\\
\hline\hline
\multirow{4}{*}{Privacy risk} &Unconstrained&0.726&0.711&0.541&0.510\\
&Fair ($\delta=0.1$)&0.744&0.777&0.550&0.513\\
&Fair ($\delta=0.01$)&0.743&0.810&0.553&0.514\\
&Fair ($\delta=0.001$)&0.745&0.818&0.555&0.515\\
\bottomrule
\end{tabularx}
\label{tab:different_delta}
\end{table}
}



{\renewcommand{\arraystretch}{1.15}
\begin{table}[t!]
\caption{\small Privacy risk of unconstrained and fair models on Law (race) dataset (with $\delta = 0.001$) -- Decision tree models. The ``DT-$x$'' row shows the results on decision tree models with max depth $x$.
}
\begin{tabularx}{\columnwidth}{
c
c
>{\centering\arraybackslash}X c
>{\centering\arraybackslash}X c
>{\centering\arraybackslash}X c
>{\centering\arraybackslash}X c
}
\toprule
 Model type & Model & $G_{0}^-$& ${G_{1}^-}^*$ & $G_{0}^+$ &$G_{1}^+$  \\ \hline
\multirow{2}{*}{DT-5}&
Unconstrained&0.585&0.561&0.515&0.503\\
&Fair&0.574&0.583&0.517&0.504\\
\hline\hline
\multirow{2}{*}{DT-10}&
Unconstrained&0.726&0.711&0.541&0.510\\
&Fair&0.745&0.818&0.555&0.515\\
\hline\hline
\multirow{2}{*}{DT -15}&
Unconstrained&0.815&0.874&0.557&0.516\\
&Fair&0.879&0.955&0.589&0.527\\
\bottomrule
\end{tabularx}
\label{tab:model_complexity_DT}
\end{table}
}

\paragraphbe{Effect of model complexity on privacy cost of fairness}
Table~\ref{tab:model_complexity_DT} shows the privacy risk across different subgroups, when we change the complexity of the decision tree models (by controlling their maximum depth).  When the model has a lower complexity, i.e., the max depth of the decision tree model is low, the privacy cost is negligible. This is because the fair models have a low accuracy even on the training dataset.  For the decision tree models with max depth $5$, after imposing fairness constraints, the test accuracy drops from $33.4\%$ to $13.3\%$ on subgroup $G_{0}^-$, and increases from $9.2\%$ only to $13.1\%$ on $G_{1}^-$.  Note that the random guessing accuracy is $50\%$.  Therefore, when the model has a lower capacity, the model cannot fit the data well and the privacy risk is low, for both fair and unconstrained model.


\section{Related Work}\label{sec:related-work}

\subsection{Algorithmic Fairness}

Various notions of algorithmic fairness are studied in the literature.  These include metric equality across sensitive groups~\cite{hardt2016equality, calders2009building}, individual fairness~\cite{dwork2012fairness}, and causality-based measures~\cite{kusner2017counterfactual}.  For training models that satisfy these fairness definitions, many techniques are proposed in the recent years, which include pre-processing methods~\cite{zemel2013learning, madras2018learning}, in-processing methods~\cite{zhang2018mitigating, kamishima2011fairness, zafar2015fairness, zafar2017fairness, agarwal2018reductions}, and post-processing methods~\cite{hardt2016equality}.  In the pre-processing methods, the goal is to find a new representation of data to retain information of input features about the learning task while scrapping the information that can lead to bias.  In-processing methods enforce fairness during the training process, by incorporating the fairness measures into the objective function or by reducing the constrained optimization problems to a sequence of cost-sensitive classification problems.  Post-processing methods correct a given model's predictions to satisfy fairness criteria.

\subsection{Membership Inference Attacks}

In the machine learning context, the membership inference
attacks aim to determine whether a given data point has been part of a model's training set~\cite{shokri2017membership, salem2018ml, nasr2019comprehensive, yeom2018privacy}.  Membership inference attacks are used to measure the information leakage of machine learning algorithms about the individual data records in their training set.  This approach is used on classification models~\cite{shokri2017membership}, adversarially robust learning algorithms~\cite{song2019privacy}, model explanations~\cite{shokri2019privacy}, embedding algorithms~\cite{song2020information} and reinforcement learning algorithms~\cite{pan2019you}.   

\subsection{Privacy and Fairness} 

In decision-making processes where fairness is a pressing need, the training dataset typically contains sensitive information about individuals (e.g., in the case of loan approval application, health condition assessment, and the recidivism prediction which we describe in Section~\ref{sec:introduction}).  

\citet{shokri2017membership} and \citet{yaghini2019disparate} demonstrate the information leakage of classification models about their training data varies across different classes and groups.  Imposing fairness constraints during the training does not eliminate the disparity of vulnerability.  The follow-up work shows that the privacy risk of some data records can be notably high even when the average privacy risk is low~\cite{longpragmatic}.  A recent interesting work studies the effect of the balance of the dataset on the privacy risk of the standard (unconstrained) models~\cite{tonni2020data}.  These results are consistent with the findings we have on unconstrained models. However, in this paper, we focus mainly on the effect of fairness constraints on the privacy risk of subgroups and individuals, instead of only analyzing the disparity of privacy loss across the population. 

\citet{dwork2012fairness} explore the relationship between fairness in machine learning and differential privacy. The authors point out that differential privacy tools can be adopted for satisfying fairness constraints. Later on, \citet{ekstrand2018privacy} pose multiple questions regarding the relation between fairness and privacy. Our results provide answers to one of the questions, and show that fairness could reduce the privacy of its subjects. 

\citet{kuppam2019fair} show that resource allocation, based on differentially private statistics, can disproportionately affect some subgroups.  In the machine learning context, \citet{bagdasaryan2019differential} study the impact of differential privacy on the prediction accuracy on subgroups, and demonstrate that if the original (unconstrained) model is unfair, the unfairness becomes worse once privacy-preserving algorithms (i.e., DP-SGD~\cite{abadi2016deep}) are applied.

\subsection{Fair Learning with Differential Privacy} 

A rigorous framework to protect privacy is to use differentially private training algorithms to bound the model's information leakage about the members of its training set.  For a given false positive rate, the true positive rate of the membership inference attacks is upper-bounded in differentially private models (as also shown in~\cite[Proposition-1]{erlingsson2019we}). Thus, the privacy risk (in Definition~\ref{def:membership_advantage}) is upper bounded by $(e^\epsilon+\delta)/2$ when the model is $(\epsilon,\delta)$-differentially private (for all the individuals regardless of their subgroups). Several works propose algorithms to learn a model that satisfies differential privacy and fairness (including demographic parity~\cite{ding2020differentially, xu2019achieving}, equality of opportunity~\cite{cummings2019compatibility}, and equalized odds~\cite{jagielski2019differentially}). Recently, \citet{tran2020differentially} introduce a differential privacy framework to train deep learning models that satisfy several group fairness notions, including equalized odds, accuracy parity, and demographic parity. 

It is important to highlight that the differential privacy considered in  \cite{jagielski2019differentially, tran2020differentially} is to protect privacy with respect to the sensitive attribute instead of the membership. \citet{jagielski2019differentially} present two algorithms that can achieve fairness and differential privacy with respect to the protected attribute. The major observation, however, is that model accuracy significantly drops.  

On the contrary, we analyze the privacy risks of individual data records and subgroups when enforcing fairness constraints.  \citet{cummings2019compatibility} present the most related theoretical results by showing the impossibility of achieving pure differential privacy and exact (equality of opportunity) fairness. 


\section{Conclusions}

In this paper, we have presented a simple yet effective framework for analyzing privacy risks of group fairness algorithms for machine learning.  We have shown that fair algorithms tend to memorize data from the under-represented subgroups, while trying to equalize the model's error across groups (partitioned based on their protected attribute).  This memorization leads to an increase in the model's information leakage about unprivileged groups.  We have provided comprehensive evaluations (using membership inference attacks) on synthetic data, as well as real data, to show how and why fair models leak information about their training data.

\section*{Acknowledgments}

The authors would like to thank Ergute Bao, Anmin Kang, Sasi Kumar Murakonda, Ta Duy Nguyen, and Martin Strobel for helpful discussions and their feedback.  This work is supported in part by the Singapore Ministry of Education Academic Research Fund, R-252-000-660-133, the NUS Early Career Research Award (NUS ECRA), grant number NUS ECRAFY19 P16, the National Research Foundation, Singapore under its Strategic Capability Research Centres Funding Initiative, and Intel within the www.private-ai.org center.

\bibliographystyle{IEEEtranN}
\bibliography{reference}

\begin{thebibliography}{45}
\providecommand{\natexlab}[1]{#1}
\providecommand{\url}[1]{#1}
\csname url@samestyle\endcsname
\providecommand{\newblock}{\relax}
\providecommand{\bibinfo}[2]{#2}
\providecommand{\BIBentrySTDinterwordspacing}{\spaceskip=0pt\relax}
\providecommand{\BIBentryALTinterwordstretchfactor}{4}
\providecommand{\BIBentryALTinterwordspacing}{\spaceskip=\fontdimen2\font plus
\BIBentryALTinterwordstretchfactor\fontdimen3\font minus
  \fontdimen4\font\relax}
\providecommand{\BIBforeignlanguage}[2]{{%
\expandafter\ifx\csname l@#1\endcsname\relax
\typeout{** WARNING: IEEEtranN.bst: No hyphenation pattern has been}%
\typeout{** loaded for the language `#1'. Using the pattern for}%
\typeout{** the default language instead.}%
\else
\language=\csname l@#1\endcsname
\fi
#2}}
\providecommand{\BIBdecl}{\relax}
\BIBdecl

\bibitem[Buolamwini and Gebru(2018)]{buolamwini2018gender}
J.~Buolamwini and T.~Gebru, ``Gender shades: Intersectional accuracy
  disparities in commercial gender classification,'' in \emph{Conference on
  fairness, accountability and transparency}, 2018, pp. 77--91.

\bibitem[Bolukbasi et~al.(2016)Bolukbasi, Chang, Zou, Saligrama, and
  Kalai]{bolukbasi2016man}
T.~Bolukbasi, K.-W. Chang, J.~Y. Zou, V.~Saligrama, and A.~T. Kalai, ``Man is
  to computer programmer as woman is to homemaker? debiasing word embeddings,''
  in \emph{Advances in neural information processing systems}, 2016, pp.
  4349--4357.

\bibitem[Agarwal et~al.(2018)Agarwal, Beygelzimer, Dud{\'\i}k, Langford, and
  Wallach]{agarwal2018reductions}
A.~Agarwal, A.~Beygelzimer, M.~Dud{\'\i}k, J.~Langford, and H.~Wallach, ``A
  reductions approach to fair classification,'' \emph{arXiv preprint
  arXiv:1803.02453}, 2018.

\bibitem[Calders et~al.(2009)Calders, Kamiran, and
  Pechenizkiy]{calders2009building}
T.~Calders, F.~Kamiran, and M.~Pechenizkiy, ``Building classifiers with
  independency constraints,'' in \emph{2009 IEEE International Conference on
  Data Mining Workshops}.\hskip 1em plus 0.5em minus 0.4em\relax IEEE, 2009,
  pp. 13--18.

\bibitem[Dwork et~al.(2012)Dwork, Hardt, Pitassi, Reingold, and
  Zemel]{dwork2012fairness}
C.~Dwork, M.~Hardt, T.~Pitassi, O.~Reingold, and R.~Zemel, ``Fairness through
  awareness,'' in \emph{Proceedings of the 3rd innovations in theoretical
  computer science conference}, 2012, pp. 214--226.

\bibitem[Hardt et~al.(2016)Hardt, Price, and Srebro]{hardt2016equality}
M.~Hardt, E.~Price, and N.~Srebro, ``Equality of opportunity in supervised
  learning,'' in \emph{Advances in neural information processing systems},
  2016, pp. 3315--3323.

\bibitem[Kamishima et~al.(2011)Kamishima, Akaho, and
  Sakuma]{kamishima2011fairness}
T.~Kamishima, S.~Akaho, and J.~Sakuma, ``Fairness-aware learning through
  regularization approach,'' in \emph{2011 IEEE 11th International Conference
  on Data Mining Workshops}.\hskip 1em plus 0.5em minus 0.4em\relax IEEE, 2011,
  pp. 643--650.

\bibitem[Kusner et~al.(2017)Kusner, Loftus, Russell, and
  Silva]{kusner2017counterfactual}
M.~J. Kusner, J.~Loftus, C.~Russell, and R.~Silva, ``Counterfactual fairness,''
  in \emph{Advances in neural information processing systems}, 2017, pp.
  4066--4076.

\bibitem[Madras et~al.(2018)Madras, Creager, Pitassi, and
  Zemel]{madras2018learning}
D.~Madras, E.~Creager, T.~Pitassi, and R.~Zemel, ``Learning adversarially fair
  and transferable representations,'' \emph{arXiv preprint arXiv:1802.06309},
  2018.

\bibitem[Zafar et~al.(2015)Zafar, Valera, Rodriguez, and
  Gummadi]{zafar2015fairness}
M.~B. Zafar, I.~Valera, M.~G. Rodriguez, and K.~P. Gummadi, ``Fairness
  constraints: Mechanisms for fair classification,'' \emph{arXiv preprint
  arXiv:1507.05259}, 2015.

\bibitem[Zafar et~al.(2017)Zafar, Valera, Rogriguez, and
  Gummadi]{zafar2017fairness}
M.~B. Zafar, I.~Valera, M.~G. Rogriguez, and K.~P. Gummadi, ``Fairness
  constraints: Mechanisms for fair classification,'' in \emph{Artificial
  Intelligence and Statistics}.\hskip 1em plus 0.5em minus 0.4em\relax PMLR,
  2017, pp. 962--970.

\bibitem[Zemel et~al.(2013)Zemel, Wu, Swersky, Pitassi, and
  Dwork]{zemel2013learning}
R.~Zemel, Y.~Wu, K.~Swersky, T.~Pitassi, and C.~Dwork, ``Learning fair
  representations,'' in \emph{International Conference on Machine Learning},
  2013, pp. 325--333.

\bibitem[Song and Raghunathan(2020)]{song2020information}
C.~Song and A.~Raghunathan, ``Information leakage in embedding models,''
  \emph{arXiv preprint arXiv:2004.00053}, 2020.

\bibitem[Shokri et~al.(2019)Shokri, Strobel, and Zick]{shokri2019privacy}
R.~Shokri, M.~Strobel, and Y.~Zick, ``Privacy risks of explaining machine
  learning models,'' \emph{arXiv preprint arXiv:1907.00164}, 2019.

\bibitem[Yeom et~al.(2018)Yeom, Giacomelli, Fredrikson, and
  Jha]{yeom2018privacy}
S.~Yeom, I.~Giacomelli, M.~Fredrikson, and S.~Jha, ``Privacy risk in machine
  learning: Analyzing the connection to overfitting,'' in \emph{2018 IEEE 31st
  Computer Security Foundations Symposium (CSF)}.\hskip 1em plus 0.5em minus
  0.4em\relax IEEE, 2018, pp. 268--282.

\bibitem[Song et~al.(2019)Song, Shokri, and Mittal]{song2019privacy}
L.~Song, R.~Shokri, and P.~Mittal, ``Privacy risks of securing machine learning
  models against adversarial examples,'' in \emph{Proceedings of the 2019 ACM
  SIGSAC Conference on Computer and Communications Security}, 2019, pp.
  241--257.

\bibitem[Shokri et~al.(2017)Shokri, Stronati, Song, and
  Shmatikov]{shokri2017membership}
R.~Shokri, M.~Stronati, C.~Song, and V.~Shmatikov, ``Membership inference
  attacks against machine learning models,'' in \emph{2017 IEEE Symposium on
  Security and Privacy (SP)}.\hskip 1em plus 0.5em minus 0.4em\relax IEEE,
  2017, pp. 3--18.

\bibitem[Sablayrolles et~al.(2019)Sablayrolles, Douze, Schmid, Ollivier, and
  J{\'e}gou]{sablayrolles2019white}
A.~Sablayrolles, M.~Douze, C.~Schmid, Y.~Ollivier, and H.~J{\'e}gou,
  ``White-box vs black-box: Bayes optimal strategies for membership
  inference,'' in \emph{International Conference on Machine Learning}, 2019,
  pp. 5558--5567.

\bibitem[Nasr et~al.(2019)Nasr, Shokri, and Houmansadr]{nasr2019comprehensive}
M.~Nasr, R.~Shokri, and A.~Houmansadr, ``Comprehensive privacy analysis of deep
  learning: Passive and active white-box inference attacks against centralized
  and federated learning,'' in \emph{2019 IEEE Symposium on Security and
  Privacy (SP)}.\hskip 1em plus 0.5em minus 0.4em\relax IEEE, 2019, pp.
  739--753.

\bibitem[Abadi et~al.(2016)Abadi, Chu, Goodfellow, McMahan, Mironov, Talwar,
  and Zhang]{abadi2016deep}
M.~Abadi, A.~Chu, I.~Goodfellow, H.~B. McMahan, I.~Mironov, K.~Talwar, and
  L.~Zhang, ``Deep learning with differential privacy,'' in \emph{Proceedings
  of the 2016 ACM SIGSAC Conference on Computer and Communications Security},
  2016, pp. 308--318.

\bibitem[Cummings et~al.(2019)Cummings, Gupta, Kimpara, and
  Morgenstern]{cummings2019compatibility}
R.~Cummings, V.~Gupta, D.~Kimpara, and J.~Morgenstern, ``On the compatibility
  of privacy and fairness,'' in \emph{Adjunct Publication of the 27th
  Conference on User Modeling, Adaptation and Personalization}, 2019, pp.
  309--315.

\bibitem[Chaudhuri et~al.(2011)Chaudhuri, Monteleoni, and
  Sarwate]{chaudhuri2011differentially}
K.~Chaudhuri, C.~Monteleoni, and A.~D. Sarwate, ``Differentially private
  empirical risk minimization,'' \emph{Journal of Machine Learning Research},
  vol.~12, no. Mar, pp. 1069--1109, 2011.

\bibitem[Papernot et~al.(2018)Papernot, Song, Mironov, Raghunathan, Talwar, and
  Erlingsson]{papernot2018scalable}
N.~Papernot, S.~Song, I.~Mironov, A.~Raghunathan, K.~Talwar, and
  {\'U}.~Erlingsson, ``Scalable private learning with pate,'' \emph{arXiv
  preprint arXiv:1802.08908}, 2018.

\bibitem[Bagdasaryan et~al.(2019)Bagdasaryan, Poursaeed, and
  Shmatikov]{bagdasaryan2019differential}
E.~Bagdasaryan, O.~Poursaeed, and V.~Shmatikov, ``Differential privacy has
  disparate impact on model accuracy,'' in \emph{Advances in Neural Information
  Processing Systems}, 2019, pp. 15\,479--15\,488.

\bibitem[Ding et~al.(2020)Ding, Zhang, Li, Wang, Yu, and
  Pan]{ding2020differentially}
J.~Ding, X.~Zhang, X.~Li, J.~Wang, R.~Yu, and M.~Pan, ``Differentially private
  and fair classification via calibrated functional mechanism,'' in
  \emph{Proceedings of the AAAI Conference on Artificial Intelligence},
  vol.~34, no.~01, 2020, pp. 622--629.

\bibitem[Xu et~al.(2019)Xu, Yuan, and Wu]{xu2019achieving}
D.~Xu, S.~Yuan, and X.~Wu, ``Achieving differential privacy and fairness in
  logistic regression,'' in \emph{Companion Proceedings of The 2019 World Wide
  Web Conference}, 2019, pp. 594--599.

\bibitem[Wightman and Ramsey(1998)]{lawdataset}
L.~F. Wightman and H.~Ramsey, ``Law school admission council.'' 1998.

\bibitem[Dua and Graff(2017)]{ucidataset}
\BIBentryALTinterwordspacing
D.~Dua and C.~Graff, ``{UCI} machine learning repository,'' 2017. [Online].
  Available: \url{http://archive.ics.uci.edu/ml}
\BIBentrySTDinterwordspacing

\bibitem[Larson et~al.(2017)Larson, Mattu, Kirchner, and Angwin]{compasdataset}
J.~Larson, S.~Mattu, L.~Kirchner, and J.~Angwin, ``{COMPAS dataset},''
  \url{https://github.com/propublica/compas-analysis}, 2017, [COMPAS dataset
  (2017)].

\bibitem[Donini et~al.(2018)Donini, Oneto, Ben-David, Shawe-Taylor, and
  Pontil]{donini2018empirical}
M.~Donini, L.~Oneto, S.~Ben-David, J.~S. Shawe-Taylor, and M.~Pontil,
  ``Empirical risk minimization under fairness constraints,'' in \emph{Advances
  in Neural Information Processing Systems}, 2018, pp. 2791--2801.

\bibitem[Dwork et~al.(2006)Dwork, McSherry, Nissim, and
  Smith]{dwork2006calibrating}
C.~Dwork, F.~McSherry, K.~Nissim, and A.~Smith, ``Calibrating noise to
  sensitivity in private data analysis,'' in \emph{Theory of cryptography
  conference}.\hskip 1em plus 0.5em minus 0.4em\relax Springer, 2006, pp.
  265--284.

\bibitem[Pan et~al.(2019)Pan, Wang, Zhang, Li, Yi, and Song]{pan2019you}
X.~Pan, W.~Wang, X.~Zhang, B.~Li, J.~Yi, and D.~Song, ``How you act tells a
  lot: Privacy-leaking attack on deep reinforcement learning,'' in
  \emph{Proceedings of the 18th International Conference on Autonomous Agents
  and MultiAgent Systems}, 2019, pp. 368--376.

\bibitem[Zafar et~al.(2019)Zafar, Valera, Gomez-Rodriguez, and
  Gummadi]{zafar2019fairness}
M.~B. Zafar, I.~Valera, M.~Gomez-Rodriguez, and K.~P. Gummadi, ``Fairness
  constraints: A flexible approach for fair classification.'' \emph{J. Mach.
  Learn. Res.}, vol.~20, no.~75, pp. 1--42, 2019.

\bibitem[Feldman(2020)]{feldman2020does}
V.~Feldman, ``Does learning require memorization? a short tale about a long
  tail,'' in \emph{Proceedings of the 52nd Annual ACM SIGACT Symposium on
  Theory of Computing}, 2020, pp. 954--959.

\bibitem[Bellamy et~al.(2018)Bellamy, Dey, Hind, Hoffman, Houde, Kannan, Lohia,
  Martino, Mehta, Mojsilovic, et~al.]{bellamy2018ai}
R.~K. Bellamy, K.~Dey, M.~Hind, S.~C. Hoffman, S.~Houde, K.~Kannan, P.~Lohia,
  J.~Martino, S.~Mehta, A.~Mojsilovic \emph{et~al.}, ``Ai fairness 360: An
  extensible toolkit for detecting, understanding, and mitigating unwanted
  algorithmic bias,'' \emph{arXiv preprint arXiv:1810.01943}, 2018.

\bibitem[Zhang et~al.(2018)Zhang, Lemoine, and Mitchell]{zhang2018mitigating}
B.~H. Zhang, B.~Lemoine, and M.~Mitchell, ``Mitigating unwanted biases with
  adversarial learning,'' in \emph{Proceedings of the 2018 AAAI/ACM Conference
  on AI, Ethics, and Society}, 2018, pp. 335--340.

\bibitem[Salem et~al.(2018)Salem, Zhang, Humbert, Berrang, Fritz, and
  Backes]{salem2018ml}
A.~Salem, Y.~Zhang, M.~Humbert, P.~Berrang, M.~Fritz, and M.~Backes,
  ``Ml-leaks: Model and data independent membership inference attacks and
  defenses on machine learning models,'' \emph{arXiv preprint
  arXiv:1806.01246}, 2018.

\bibitem[Yaghini et~al.(2019)Yaghini, Kulynych, and
  Troncoso]{yaghini2019disparate}
M.~Yaghini, B.~Kulynych, and C.~Troncoso, ``Disparate vulnerability: On the
  unfairness of privacy attacks against machine learning,'' \emph{arXiv
  preprint arXiv:1906.00389}, 2019.

\bibitem[Long et~al.()Long, Wang, Bu, Bindschaedler, Wang, Tang, Gunter, and
  Chen]{longpragmatic}
Y.~Long, L.~Wang, D.~Bu, V.~Bindschaedler, X.~Wang, H.~Tang, C.~A. Gunter, and
  K.~Chen, ``A pragmatic approach to membership inferences on machine learning
  models.''

\bibitem[Tonni et~al.(2020)Tonni, Vatsalan, Farokhi, Kaafar, Lu, and
  Tangari]{tonni2020data}
S.~M. Tonni, D.~Vatsalan, F.~Farokhi, D.~Kaafar, Z.~Lu, and G.~Tangari, ``Data
  and model dependencies of membership inference attack,'' \emph{arXiv preprint
  arXiv:2002.06856}, 2020.

\bibitem[Ekstrand et~al.(2018)Ekstrand, Joshaghani, and
  Mehrpouyan]{ekstrand2018privacy}
M.~D. Ekstrand, R.~Joshaghani, and H.~Mehrpouyan, ``Privacy for all: Ensuring
  fair and equitable privacy protections,'' in \emph{Conference on Fairness,
  Accountability and Transparency}, 2018, pp. 35--47.

\bibitem[Kuppam et~al.(2019)Kuppam, McKenna, Pujol, Hay, Machanavajjhala, and
  Miklau]{kuppam2019fair}
S.~Kuppam, R.~McKenna, D.~Pujol, M.~Hay, A.~Machanavajjhala, and G.~Miklau,
  ``Fair decision making using privacy-protected data,'' \emph{arXiv preprint
  arXiv:1905.12744}, 2019.

\bibitem[Erlingsson et~al.(2019)Erlingsson, Mironov, Raghunathan, and
  Song]{erlingsson2019we}
{\'U}.~Erlingsson, I.~Mironov, A.~Raghunathan, and S.~Song, ``That which we
  call private,'' \emph{arXiv preprint arXiv:1908.03566}, 2019.

\bibitem[Jagielski et~al.(2019)Jagielski, Kearns, Mao, Oprea, Roth,
  Sharifi-Malvajerdi, and Ullman]{jagielski2019differentially}
M.~Jagielski, M.~Kearns, J.~Mao, A.~Oprea, A.~Roth, S.~Sharifi-Malvajerdi, and
  J.~Ullman, ``Differentially private fair learning,'' in \emph{International
  Conference on Machine Learning}.\hskip 1em plus 0.5em minus 0.4em\relax PMLR,
  2019, pp. 3000--3008.

\bibitem[Tran et~al.(2020)Tran, Fioretto, and
  Van~Hentenryck]{tran2020differentially}
C.~Tran, F.~Fioretto, and P.~Van~Hentenryck, ``Differentially private and fair
  deep learning: A lagrangian dual approach,'' \emph{arXiv preprint
  arXiv:2009.12562}, 2020.

\end{thebibliography}
\newpage
\appendices


\section{Additional Results}\label{appendix-additional-results}

Table~\ref{tab:accuracy_subgroups_real_data} shows the prediction accuracy of unconstrained and fair models on all subgroups for decision tree models with max depth $10$.

{
\begin{table}[h!]
\caption{\small Prediction accuracy, decision tree (max depth $10$). }
\centering
\begin{tabularx}{1.6\columnwidth}{
c
>{\centering\arraybackslash}X c
>{\centering\arraybackslash}X c
>{\centering\arraybackslash}X c
>{\centering\arraybackslash}X c
>{\centering\arraybackslash}X c
>{\centering\arraybackslash}X c
}
\toprule
Dataset& Model &  &$G_{0}^-$& $G_{1}^-$ & $G_{0}^+$ &$G_{1}^+$  \\ \hline
\multirow{2}{*}{Bank}
&\multirow{2}{*}{Unconstrained}
&Train acc&95.7\%&97.7\%&79.8\%&74.3\%\\

&&Test acc &88.5\%&94.5\%&56.1\%&53.3\%\\

\cline{2-7}
\multirow{2}{*}{(age)}&\multirow{2}{*}{Fair ($\delta=0.001$)}
&Train acc&97.3\%&97.3\%&75.3\%&75.2\%\\

&&Test acc &89.6\%&94.4\%&51.1\%&54.6\%\\

\hline\hline

\multirow{2}{*}{COMPAS }
&\multirow{2}{*}{Unconstrained}
&Train acc&81.9\%&92.3\%&74.2\%&61.4\%\\

&&Test acc &67.8\%&81.1\%&60.0\%&42.2\%\\

\cline{2-7}
\multirow{2}{*}{(race)}&\multirow{2}{*}{Fair ($\delta=0.001$)}
&Train acc&85.6\%&85.7\%&70.8\%&70.8\%\\

&&Test acc &70.9\%&73.4\%&56.8\%&50.3\%\\

\hline\hline

\multirow{2}{*}{COMPAS}
&\multirow{2}{*}{Unconstrained}
&Train acc&92.0\%&82.5\%&72.7\%&71.6\%\\

&&Test acc &79.2\%&70.6\%&47.0\%&57.8\%\\

\cline{2-7}

\multirow{2}{*}{(gender)}&\multirow{2}{*}{Fair ($\delta=0.001$)}
&Train acc&84.6\%&84.4\%&71.6\%&71.5\%\\

&&Test acc &73.9\%&70.8\%&49.9\%&57.2\%\\

\hline\hline

\multirow{2}{*}{Law}
&\multirow{2}{*}{Unconstrained}
&Train acc&70.1\%&43.4\%&99.0\%&99.8\%\\

&&Test acc &28.6\%&10.8\%&92.2\%&98.6\%\\

\cline{2-7}
\multirow{2}{*}{(race)}&\multirow{2}{*}{Fair ($\delta=0.001$)}
&Train acc&54.5\%&54.3\%&99.6\%&99.7\%\\

&&Test acc &23.3\%&11.8\%&94.0\%&98.1\%\\

\hline\hline

\multirow{2}{*}{Law}
&\multirow{2}{*}{Unconstrained}
&Train acc&57.3\%&54.2\%&99.6\%&99.7\%\\
&&Test acc &20.8\%&18.0\%&97.4\%&97.8\%\\
\cline{2-7}
\multirow{2}{*}{(gender)}&\multirow{2}{*}{Fair ($\delta=0.001$)}
&Train acc&57.2\%&57.1\%&99.8\%&99.7\%\\
&&Test acc &20.7\%&17.8\%&97.4\%&97.7\%\\

\bottomrule
\end{tabularx}
\label{tab:accuracy_subgroups_real_data}
\end{table}
}

\end{document}